\documentclass[10pt,twocolumn,letterpaper]{article}

\usepackage{iccv}
\usepackage{times}
\usepackage{epsfig}
\usepackage{graphicx}
\usepackage{amsmath}
\usepackage{amssymb}

\usepackage[tableposition=above]{caption}
\usepackage{subcaption}
\usepackage{bm}
\usepackage{bbm}
\usepackage{xspace}
\usepackage[utf8]{inputenc} 
\usepackage{url}            
\usepackage{booktabs}       
\usepackage{amsfonts}       
\usepackage{nicefrac}       
\usepackage{microtype}      
\usepackage{color}
\usepackage{algorithm}
\usepackage{algorithmic}
\usepackage{array}
\usepackage{multirow}
\usepackage{float}
\usepackage{enumitem}
\usepackage{makecell}
\usepackage{mathtools}
\usepackage[permil]{overpic}
\newcommand{\iti}{\textit{i})\@\xspace}
\newcommand{\itii}{\textit{ii})\@\xspace}
\newcommand{\itiii}{\textit{iii})\@\xspace}

\definecolor{red}{rgb}{1,0,0}
\definecolor{green}{rgb}{0,0.6,0}
\definecolor{blue}{rgb}{0,0,1}
\definecolor{cyan}{rgb}{0,0.5,0.5}
\definecolor{orange}{rgb}{0.9,0.4,0}
\definecolor{slateblue}{rgb}{0.7,0.35,0.9}
\definecolor{brown}{rgb}{0.3, 0.2, 0}
\definecolor{mahogany}{rgb}{0.75, 0.25, 0.0}
\definecolor{purple}{rgb}{0.5, 0, 0.5}
\definecolor{darkgreen}{rgb}{0, 0.4, 0}
\definecolor{frenchblue}{rgb}{0.0, 0.45, 0.73}
\definecolor{goldenrod}{rgb}{0.65, 0.45, 0.03}
\definecolor{gray}{rgb}{0.5,0.5,0.5}

\makeatletter
\renewcommand{\paragraph}{%
  \@startsection{paragraph}{4}%
  {\z@}{0.3\baselineskip \@plus 0ex \@minus 0ex}{-1em}%
  {\normalfont\normalsize\bfseries}%
}
\makeatother

\makeatletter
\def\@fnsymbol#1{\ensuremath{\ifcase#1\or 1\or 2\or 3\or
   4\or 5\or 6\or *\or \dagger
   \or \ddagger\ddagger \else\@ctrerr\fi}}
\makeatother

\usepackage[breaklinks=true,bookmarks=false]{hyperref}

\iccvfinalcopy 

\def\iccvPaperID{****} 

\begin{document}

\title{\large Specialize and Fuse: Pyramidal Output Representation for Semantic Segmentation}

\author{
Chi-Wei Hsiao\setcounter{footnote}{6}\thanks{The authors contribute equally to this paper.}\setcounter{footnote}{0}~\footnotemark[1]
\qquad Cheng Sun\footnotemark[7]~\footnotemark[1]~$^{,}$\footnotemark[2] \qquad Hwann-Tzong Chen\footnotemark[1]~$^{,}$\footnotemark[3] \qquad Min Sun\footnotemark[1]~$^{,}$\footnotemark[4] \\
{\small National Tsing Hua University\footnotemark[1]} \qquad {\small ASUS AICS Department\footnotemark[2]} \qquad {\small Aeolus Robotics\footnotemark[3]} \\
{\small Joint Research Center for AI Technology and All Vista Healthcare\footnotemark[4]} \\
{\tt\small \{chiweihsiao,chengsun\}@gapp.nthu.edu.tw} \qquad {\tt\small htchen@cs.nthu.edu.tw} \qquad {\tt\small sunmin@ee.nthu.edu.tw}
}

\maketitle
\ificcvfinal\thispagestyle{empty}\fi

\begin{abstract}
    We present a novel pyramidal output representation to ensure parsimony with our “specialize and fuse” process for semantic segmentation.
    A pyramidal “output" representation consists of coarse-to-fine levels, where each level is “specialize” in a different class distribution (e.g., more stuff than things classes at coarser levels).
    Two types of pyramidal outputs (i.e., unity and semantic pyramid) are “fused" into the final semantic output, where the unity pyramid indicates unity-cells (i.e., all pixels in such cell share the same semantic label).
    The process ensures parsimony by predicting a relatively small number of labels for unity-cells (e.g., a large cell of grass) to build the final semantic output.
    In addition to the “output" representation, we design a coarse-to-fine contextual module to aggregate the “features" representation from different levels.
    We validate the effectiveness of each key module in our method through comprehensive ablation studies.
    Finally, our approach achieves state-of-the-art performance on three widely-used semantic segmentation datasets---ADE20K, COCO-Stuff, and Pascal-Context.
\end{abstract}

\vspace{-1em}
\section{Introduction}

Given an RGB image, semantic segmentation defines semantic labels for all pixels as ``output".
Recent methods on semantic segmentation widely exploit deep neural networks.
One major research direction is designing new contextual modules~\cite{DingJLM019,DingJSLW19,FuLT0BFL19,HuangWHHW019,YuanCW20,ZhangCLHLMHD19,ZhangDSZWTA18,ZhuXBHB19} exploring better ``feature" representation in the networks.
We argue that leveraging the structure in the ``output" representation could open up opportunities orthogonal to the current endeavors.
We observe that a large portion of pixels in most images share the same label at a coarse spatial level (\eg, stuff classes like sky and grass, or central region of objects). 
This observation induces a parsimonious strategy to dynamically predict semantic labels at a coarser level according to the spatial distribution of classes in each input image.

\begin{figure*}
  \centering
  \begin{overpic}[width=0.85\linewidth]{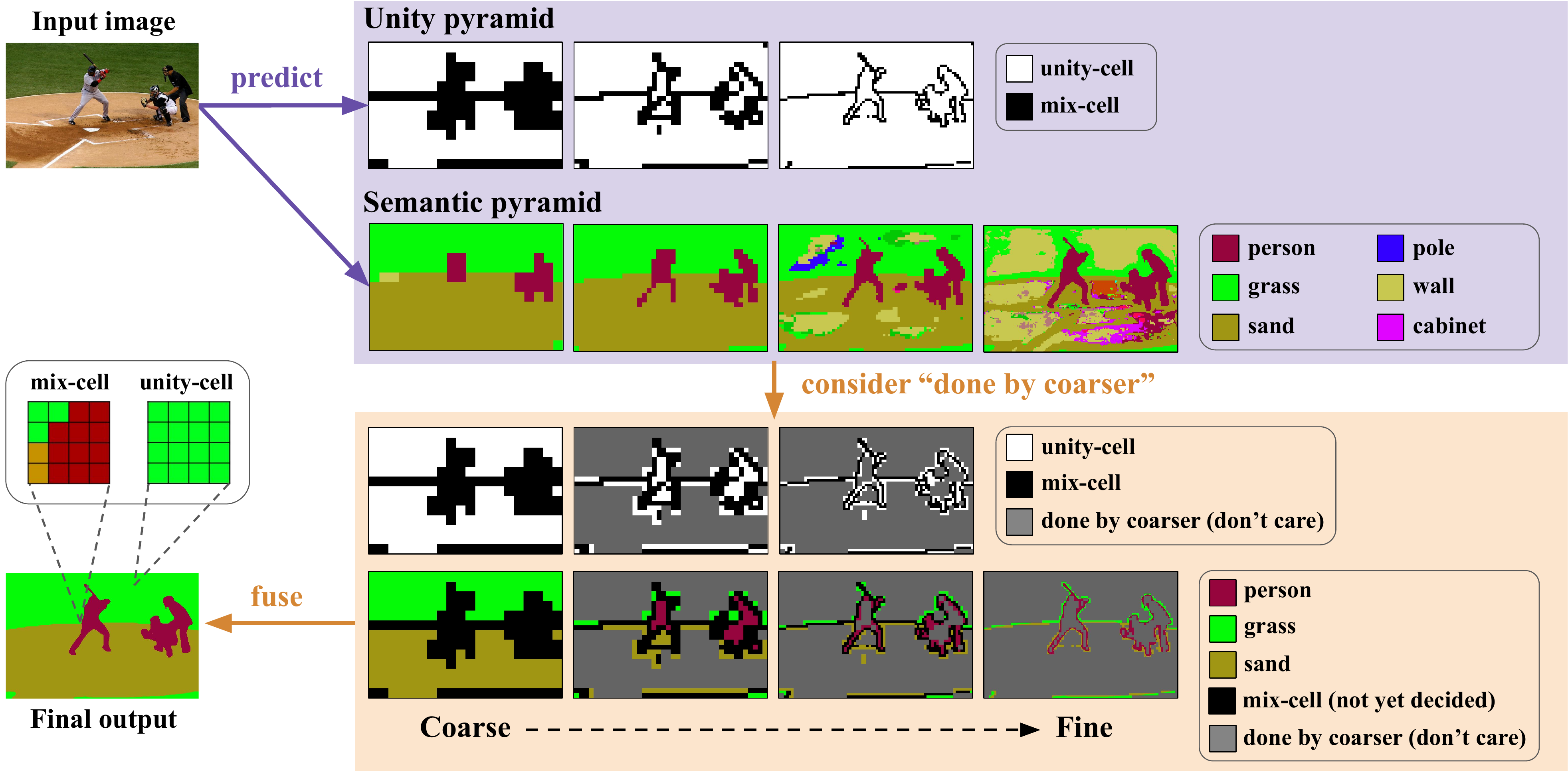}
  \put(142,68){\textcolor{orange}{\scriptsize (see Eq.~\ref{eq:fuse})}}
  \end{overpic}
  \vspace{-.5em}
  \caption{
    An overview of the ``specialize and fuse" approach. We train a neural network to predict two pyramidal outputs:
    \textbf{Unity pyramid} classifies cells into ``unity-cell" (\ie, all covered pixels share the same class) or ``mix-cell" (\ie, covered pixels contain multiple classes);
    \textbf{Semantic pyramid} predicts the semantic labels at multiple levels.
    The orange bottom panel illustrates how we fuse the two predicted pyramids into one final semantic output (the bottom-left-most image).
    Intuitively, a unity-cell at a coarser level indicates that all cells covered by it at finer levels are ``done by coarser" and thus can be ignored (colored in grey).
    Therefore, we acquire the final semantic labels from unity-cells in a coarse-to-fine manner.
    In other words, mix-cells or ``done by coarser" cells are ignored during training (Sec.~\ref{ssec:specialize}) and inference (Sec.~\ref{ssec:fuse}).
    Our approach achieves parsimony by training the network to predict a relatively small number of semantic labels for unity-cells (\ie, most cells at finer levels are ``done by coarser") and enables each pyramid level to specialize in different class distribution. 
  }
  \label{fig:overview}
  \vspace{-1em}
\end{figure*}

We proposed a novel pyramidal output representation to ensure parsimony with our ``specialize and fuse" process (Fig.~\ref{fig:overview}).
Firstly, rather than a single-level output, a pyramidal output starting from the coarsest level to the finest level is designed so that each level is learned to ``specialize" in a different class distribution (\eg, more stuff than things classes at coarser levels).
Specifically, two types of pyramidal output (unity and semantic pyramid) are predicted. Unity pyramid identifies whether a patch of pixels (referred to as a cell) shares the same label (referred to as a unity-cell) (Fig.~\ref{fig:overview}-first row), and semantic pyramid consists of semantic labels at multiple levels (Fig.~\ref{fig:overview}-second row).
Finally, the semantic pyramid are ``fused" into one single semantic output according to the unity-cells across levels (Fig.~\ref{fig:overview}-bottom panel).
Note that our ``specialize and fuse" process ensures parsimony by predicting a relatively small number of labels for unity-cells (\eg, a large cell of grass) to build the final semantic output.
In addition to the ``output" representation, we design a coarse-to-fine contextual module to aggregate the ``features'' representation from different levels for improving semantic pyramid prediction.

Our main contributions are as follows:
\iti we introduce a pyramidal ``output'' representation and a ``specialize and fuse" process to allow each level to specialize in different class distribution and ensure parsimony;
\itii we design a contextual module to aggregate the ``features'' representation from different levels for further improvements;
\itiii we showcase the effectiveness of our method on ADE20K, COCO-Stuff, and Pascal-Context. Our method with both HRNet and ResNet as the backbone can achieve results on par with or better than the recent state-of-the-art methods.

\section{Related work}

\paragraph{Contextual modules.}
Context is important to the task of semantic segmentation, with more and more improvements coming from the newly designed context spreading strategy.
PSPNet~\cite{ZhaoSQWJ17} proposes to pool deep features into several small and fixed spatial resolutions to generate global contextual information.
Deeplab~\cite{ChenZPSA18} employs dilated CNN layers with several dilation rates, which helps the model capture different ranges of context.
Recently, self-attention methods~\cite{VaswaniSPUJGKP17,WangGGH18} achieve great success in natural language processing and computer vision, with many variants being proposed for semantic segmentation.
DANet~\cite{FuLT0BFL19} applies self-attention in spatial dimension and channel dimension.
CCNet~\cite{HuangWHHW019} proposes criss-cross attention in which a pixel attends only to pixels of the same column or row.
ANL~\cite{ZhuXBHB19} pools features to a fixed spatial size, which acts as the key and value of attention.
OCR~\cite{YuanCW20} pools the context according to a coarse prediction and computes attention between deep features and class centers.
CCNet~\cite{HuangWHHW019}, ANL~\cite{ZhuXBHB19}, and OCR~\cite{YuanCW20} are able to reduce the computation via specially designed attending strategies while still retain or even improve the performance.
Inspired by ANL~\cite{ZhuXBHB19}, we design a new coarse-to-fine contextual module in this work, and furthermore, the contextual module is managed to integrate seamlessly with the proposed pyramidal output format.

\paragraph{Hierarchical semantic segmentation prediction.}
Layer Cascade (LC)~\cite{LiLLLT17} predicts three semantic maps of the same resolution using three cascaded sub-networks: Each sub-network passes uncertain pixels to the next sub-network for further prediction, and all of the LC predictions are of the same level.
In contrast, our method provides a new output representation that is independent of sub-networks in the backbone, and predicts multi-level semantic maps trained under the principle of parsimony. Besides, unlike LC which simply combines the semantic maps based on the semantic prediction itself, we train a \emph{unity pyramid} with a carefully defined physical meaning to infer with the \emph{semantic pyramid}.

PointRend~\cite{KirillovWHG20} and QGN~\cite{ChittaAH20} are two recent approaches that explore the direction of hierarchical prediction where the final semantic segmentation map is reconstructed in a coarse-to-fine manner instead of a dense prediction from the deep model directly.
Both approaches start from the coarsest prediction.
PointRend~\cite{KirillovWHG20} gradually increases the resolution by sampling only uncertain points for the finer prediction, while QGN~\cite{ChittaAH20} predicts $C$+1 classes where the extra ``composite'' class indicates whether a point would be propagated to a finer level in their SparseConv~\cite{GrahamEM18} decoder.
Both approaches yield high-resolution prediction (the same as input resolution) with their efficient sparseness design.
PointRend~\cite{KirillovWHG20} achieves slightly better mIoU by refining the prediction to a high spatial resolution; while, QGN~\cite{ChittaAH20} focuses on computational efficiency but results in inferior performance.
The hierarchical output in previous works~\cite{ChittaAH20,KirillovWHG20,LiLLLT17} entwine with their model's data flow, while our pyramidal output format is more flexible to the model architecture as long as it yields the two pyramids.
Besides, rather than results refining and computation saving, the proposed pyramidal output representation with tailored training and fusing procedures (\ie, the proposed \emph{Specialize and Fuse} strategy) is able to achieve state-of-the-art performance.

\section{Pyramidal output representation}
\label{sec:method}
The overview of our ``specialize and fuse" process given two types of pyramidal outputs are shown in Fig.~\ref{fig:overview}.
In the following, we first define the semantic and unity pyramids in Sec.~\ref{ssec:method_pyramid}. Then, the ``training to specialize" and ``fuse in inference" phases are introduced in Sec.~\ref{ssec:specialize} and Sec.~\ref{ssec:fuse}, respectively.


\subsection{Semantic pyramid and unity pyramid} \label{ssec:method_pyramid} 
\paragraph{Pyramid structure.}
We adopt the \emph{coarse-to-fine pyramid} structure to build our pyramidal output format, where a finer level has double resolution than its adjacent coarser level, and all cells (\ie, a patch of pixels) except those in the finest level have exactly four children.
Besides, the width and height of an input image should be divisible by those of the coarsest level; otherwise, we resize the input RGB to the nearest divisible.

\begin{figure*}
  \centering
  \begin{subfigure}[t]{0.39\textwidth}
      \includegraphics[width=\linewidth]{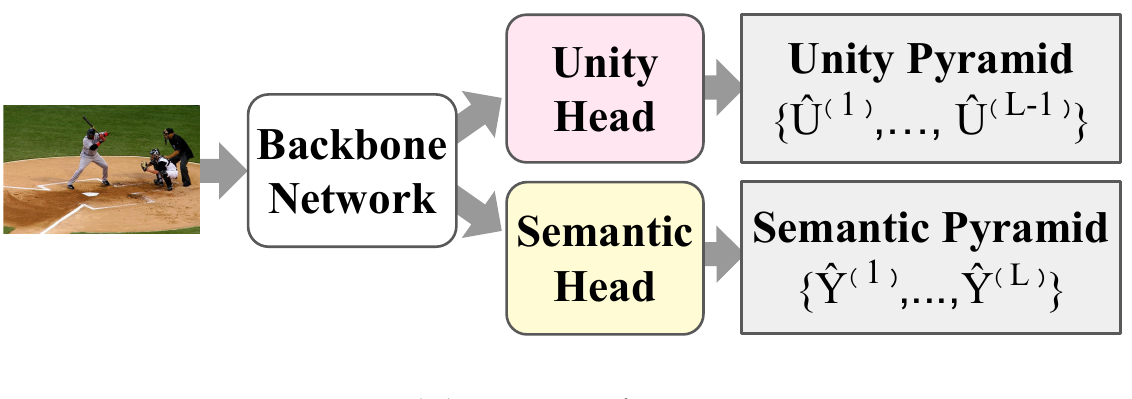}
      \caption{Overview.}
  \label{fig:arch_overview}
  \end{subfigure}
  \hfill
  \begin{subfigure}[t]{0.6\textwidth}
      \includegraphics[width=\linewidth]{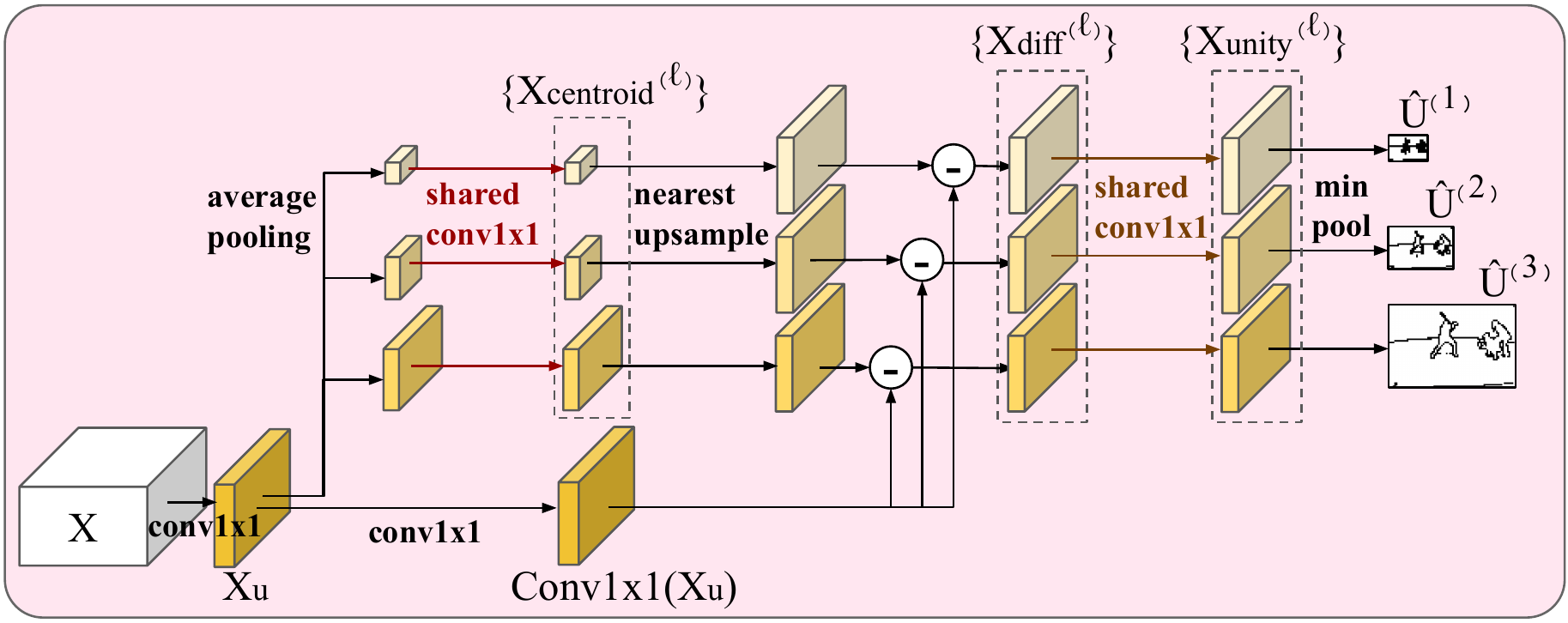}
      \caption{Unity head.}
  \label{fig:arch_uhead}
  \end{subfigure}
  \par\medskip
  \begin{subfigure}[t]{0.5216\textwidth}
      \includegraphics[width=\linewidth]{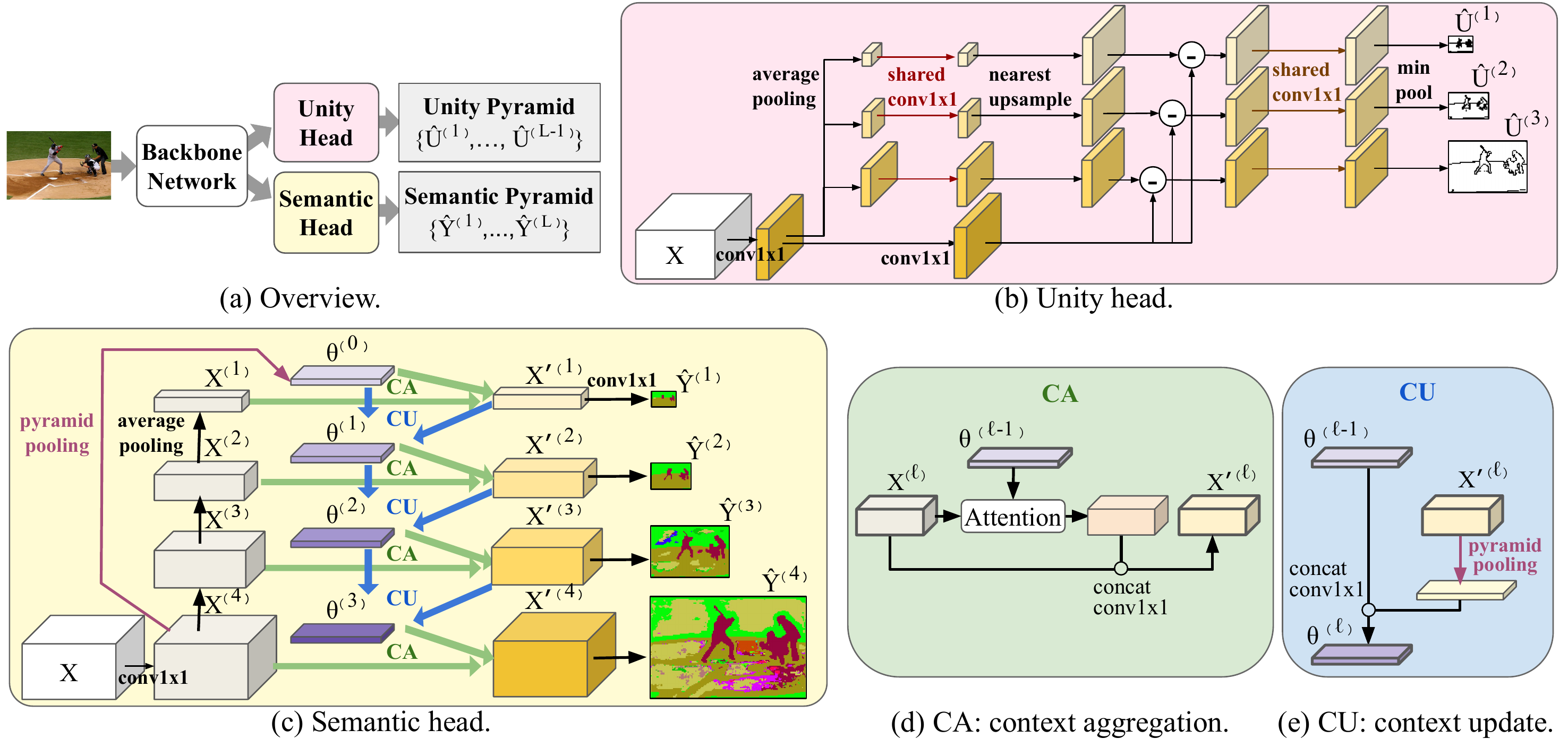}
      \caption{Semantic head.}
  \label{fig:arch_shead}
  \end{subfigure}
  \hfill
  \begin{subfigure}[t]{0.2809\textwidth}
      \includegraphics[width=\linewidth]{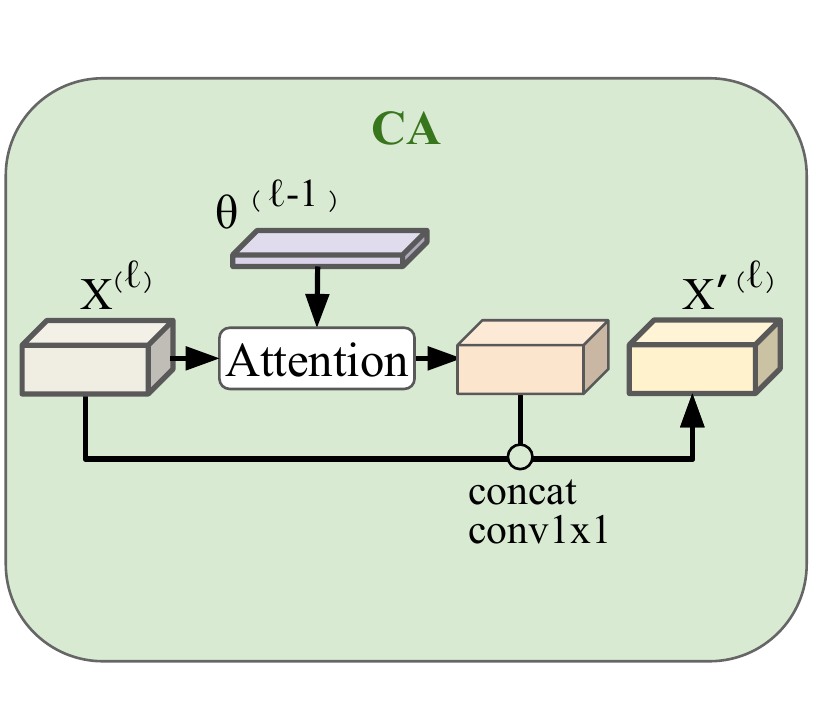}
      \caption{CA: context aggregation.}
  \label{fig:arch_ca}
  \end{subfigure}
  \hfill
  \begin{subfigure}[t]{0.1775\textwidth}
      \includegraphics[width=\linewidth]{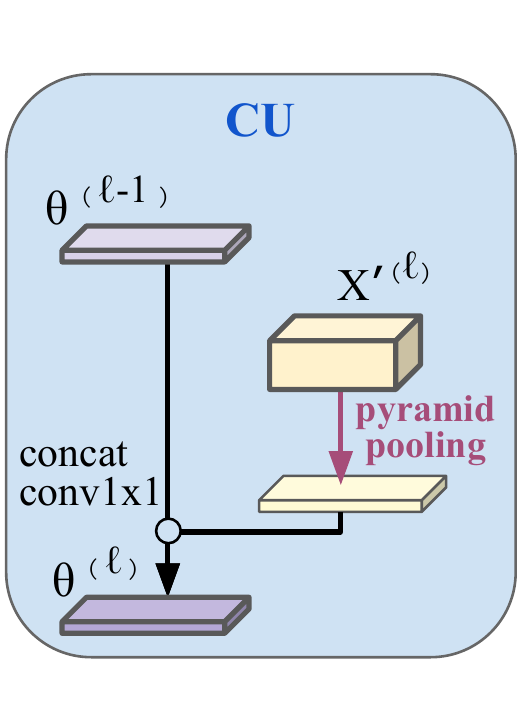}
      \caption{CU: context update.}
  \label{fig:arch_cu}
  \end{subfigure}
  \caption{
  An illustration of the neural network architecture.
  (a) Two additional heads are added to a backbone network for predicting the proposed unity pyramid and semantic pyramid.
  (b) The unity head.
  (c) For the semantic head, we design a coarse-to-fine contextual module, which comprises two operations---(d) CA: context aggregation and (e) CU: context update.
  Note that the $\hat{U}^{(\ell)}$ in (b) and  $\hat{Y}^{(\ell)}$ in (c) are raw outputs and will be fused into one to serve as the final prediction (Sec.~\ref{ssec:fuse}).}
  \label{fig:arch}
\end{figure*}

\paragraph{Notation.}
We denote the index of the pyramid level by $\ell$, where $\ell{=}1$ is the coarsest level and $\ell{=}L$ is the finest level ($L$ levels in total).
Let $s_\ell$ denote the spatial stride at the pyramid level $\ell$, $D$ the latent dimension of backbone features, and $C$ the number of output classes.
As shown in Fig.~\ref{fig:arch_uhead} and Fig.~\ref{fig:arch_shead}, our model takes a feature tensor $X \in \mathbb{R}^{D \times \frac{H}{s_L} \times \frac{W}{s_L}}$ from the backbone, and predicts a \emph{semantic pyramid} $\big\{ \hat{Y}^{(\ell)} \in \mathbb{R}^{C \times \frac{H}{s_\ell} \times \frac{W}{s_\ell}}\big\}_{\ell=1,\ldots,L}$ and a \emph{unity pyramid} $\big\{ \hat{U}^{(\ell)} \in \mathbb{R}^{\frac{H}{s_\ell} \times \frac{W}{s_\ell}}\big\}_{\ell=1,\ldots,L-1}$.
The channel dimension of $\hat{U}$ is 1 (binary classification) and is discarded.
Note that the output stride of the finest level $s_L$ is the same as the output stride of the backbone feature $X$ in this work for simplicity.
The final semantic output fused from two pyramids is denoted as $\hat{Y} \in \mathbb{R}^{C \times \frac{H}{s_L} \times \frac{W}{s_L}}$.
The pyramidal ground truths $Y^{(\ell)}, U^{(\ell)}$ for training are derived from the ground truth per-pixel semantic labeling $Y$.

\paragraph{Pyramidal ground truth.}
At pyramid level $\ell$, each cell is responsible for a patch of $s_\ell \times s_\ell$ pixels in the original image.
A cell can be either an unity cell with the same label for all pixels within the cell, or a mix-cell with more than one labels.
In the ground truth unity pyramid $U^{(\ell)}$, positive and negative values indicate unity-cells and mix-cells, respectively.
In the ground truth semantic pyramid $Y^{(\ell)}$, for a unity-cell, its ground truth semantic label is defined as the shared label by all covered pixels in the original per-pixel ground truth $Y$, whereas, for a mix-cell, its ground truth semantic label is ill-defined and ignored in computing training loss. Note that a unity-cell at level $\ell$ implies its child cells are also unity-cells at level $\ell+1$. To avoid redundancy, the children unity-cells (referred to as ``done by coarser") are ignored both in computing training loss (Sec.~\ref{ssec:specialize}) and in the fuse-phase (Sec.~\ref{ssec:fuse}).

\subsection{\emph{Specialize}---the training phase} \label{ssec:specialize}
Our experiments show that naively training the predicted $\hat{Y}^{(\ell)}, \hat{U}^{(\ell)}$ with their ground-truth counterparts $Y^{(\ell)}, U^{(\ell)}$ is unable to provide any improvement. This setting does not utilize the fact that a large number of pixels belonging to unity-cells are already predicted at a coarser level, and hence the finer level had better not be redundantly trained on those predicted regions.
Based on the motivation to encourage parsimony and to train specialized pyramid levels, for those cells whose predecessors in the pyramid structure are already correctly classified as unity-cells (true positives), our training procedure re-labels them as ``don't care'' on the fly (we refer to such labels as ``done by coarser'').
With the relabeled ground truths in each mini-batch, the training loss is computed as follows:
\begin{equation}
\begin{split}
    \mathcal{L}=&\frac{1}{L} \sum_{\ell=1}^{L} \mathrm{CE}(\hat{Y}^{(\ell)},~Y_{\mathrm{relabeled}}^{(\ell)})\\
    &+~\frac{1}{L-1} \sum_{\ell=1}^{L-1} \mathrm{BCE}(\hat{U}^{(\ell)},~U_{\mathrm{relabeled}}^{(\ell)}) \,,
\end{split}
\end{equation}
where CE is cross entropy and BCE is binary cross entropy.
Note that only $L-1$ levels are predicted in the unity pyramid $\hat{U}^{(\ell)}$ as all cells in the finest level $L$ are assumed to be unity-cells (there is no subsequent finer-level semantic prediction to be considered).
We show in the experiments that each level of the semantic pyramid has indeed learned to specialize in characterizing the assigned pixels.

\subsection{\emph{Fuse}---the inference phase} \label{ssec:fuse}
During inference, we fuse the two predicted pyramids into one final semantic map $\hat{Y}$.
Given the predicted \emph{semantic pyramid} $\hat{Y}^{(\ell)}$ and \emph{unity pyramid} $\hat{U}^{(\ell)}$, the inference procedure refers each pixel to the semantic prediction at the ``coarsest" unity-cell as follows.
\begin{equation}\label{eq:fuse}
\begin{split}
    \hat{Y} = \sum_{\ell=1}^{L} ~ &
                    \Biggl\{  \mathbbm{1}\left[\mathrm{Up}(\hat{U}^{(\ell)})\ge\tau\right] \odot
                    \mathrm{Up}(\hat{Y}^{(\ell)})
                    \\
                    & 
                    \odot   
                    \prod_{1\le k < \ell}^\odot \mathbbm{1}\left[\mathrm{Up}(\hat{U}^{(k)}) < \tau\right]
                    \Biggr\} \,,
\end{split}
\end{equation}
where $\mathrm{Up}(\cdot)$ upsamples the prediction at level $\ell$ to the finest level $L$, $\tau$ is a threshold to decide unity-cell (\ie, $\ge\tau$) or mix-cell (\ie, $<\tau$), $\odot$ and $\mathbbm{1}[\cdot]$ denote the element-wise multiplication and indicator function, respectively.
The $1^{st}$ line in Eq.~\ref{eq:fuse} selects the semantic labels at unity-cells in level $\ell$.
To ensure that the ``coarsest" unity-cell is selected, the $2^{nd}$ line in Eq.~\ref{eq:fuse} checks whether all of its preceding cells in levels $1 \sim (\ell-1)$ are mix-cells.

\section{Architectures of pyramid head} \label{sec:arch}
 The architectures of unity and semantic head are shown in Fig.~\ref{fig:arch} and detailed in the following sections.
\subsection{Unity head} \label{ssec:method_unity}
The unity head takes a feature $X$ from the backbone as input and outputs a \emph{unity pyramid} $ \hat{U}^{(\ell)}$.
The design of unity head (Fig.~\ref{fig:arch_uhead}) follows the idea that the embedding of every pixel within a unity-cell has to be close to the embeddings of the centroid of the cell as they share the same semantic class.
First, we employ a $1 \times 1$ convolution layer ($\operatorname{Conv1x1}$) to convert $X$ to $X_u$ with reduced channels $D_u$.
Next, we generate centroid embedding of all cells in a pyramid $X_{\mathrm{centroid}}^{(\ell)} \in \mathbb{R}^{D_u \times \frac{H}{s_\ell} \times \frac{W}{s_\ell}}$ by applying average pooling and a shared $\operatorname{Conv1x1}$ to $X_u$.
To measure the difference $X_{\mathrm{diff}}^{(\ell)}$ between the embedding of all pixels within a cell and the centroid embedding of the cell, we upsample each $X_{\mathrm{centroid}}^{(\ell)}$ to the finest level using nearest-neighbor interpolation and subtract it from $\operatorname{Conv1x1}(X_u)$.
Subsequently, a shared $\operatorname{Conv1x1}$ followed by a $\mathrm{sigmoid}$ function converts $X_{\mathrm{diff}}^{(\ell)} \in \mathbb{R}^{D_u \times \frac{H}{s_L} \times \frac{W}{s_L}}$ to $X_{\mathrm{unity}}^{(\ell)} \in \mathbb{R}^{\frac{H}{s_L} \times \frac{W}{s_L}}$, where each entry yields a probability that it shares the same semantic class with the cell centroid.
Reflecting the definition of a unity-cell, we use min pooling to query if the most deviated entry in a cell is similar to the cell centroid, and produce the final unity pyramid $\hat{U}^{(\ell)} \in \mathbb{R}^{\frac{H}{s_\ell} \times \frac{W}{s_\ell}}$.

\subsection{Semantic head} \label{ssec:method_module}
\paragraph{Predicting semantic in pyramidal format.}
We first employ a $\operatorname{Conv1x1}$ layer to project the number of channels from backbone's $D$ to $D_s$, producing feature $X^{(L)}$.
The simplest way to predict the semantic pyramid $\{\hat{Y}^{(\ell)}\}_{\ell=1,\ldots,L}$ is directly pooling $X^{(L)}$ to the $L$ desired spatial sizes:
\begin{equation}
    X^{(\ell)} = \operatorname{AvgPool}\left(X^{(L)}, \, 2^{L-\ell}\right) ,\label{eq:avg_pool}
\end{equation}
where $\operatorname{AvgPool}(\cdot, k)$ is average pooling with kernel size and stride set to $k$.
Each $X^{(\ell)}$ is then projected from latent dimension $D_s$ to the number of classes $C$ with convolutional layers.
We show in our experiment that such a simplest network setting can already achieve promising improvements with the proposed \emph{specialize-and-fuse} strategy.

\paragraph{Coarse-to-fine contextual module.}
Motivated by the recent success of context spreading strategy for semantic segmentation, we further design a coarse-to-fine contextual module for our pyramidal output format.
Intuitively, we aggregate the contextual information from coarser pyramid levels to assist the prediction at finer pyramid levels.
Fig.~\ref{fig:arch_shead} depicts the proposed coarse-to-fine semantic head, which refines $\{X^{(\ell)}\}_{\ell=1,\ldots,L}$ in Eq.~(\ref{eq:avg_pool}) with context feature $\{\theta^{(\ell)}\}_{\ell=0,\ldots,L-1}$ from coarse to fine level:
\begin{align}
    X'^{(\ell)} &= \operatorname{CA}^{(\ell)}\,\left(X^{(\ell)}, \theta^{(\ell-1)}\right) ,\label{eq:intro_CA} \\
    \theta^{(\ell)} &= \operatorname{CU}^{(\ell)}\,\left(X'^{(\ell)}, \theta^{(\ell-1)}\right) ,\label{eq:intro_CU}
\end{align}
where the $\operatorname{CA}$ and $\operatorname{CU}$ stand for \emph{Context Aggregation module} and \emph{Context Updating module} and will be detailed later.
Inspired by ANL's strategy ~\cite{ZhuXBHB19} to improve efficiency of contextual module, our initial context feature is
\begin{equation} \label{eq:pyramid_pool_0}
    \theta^{(0)} = \operatorname{PyramidPool}(X^{(L)}),
\end{equation}
where $\operatorname{PyramidPool}$ flatten the 1$\times$1, 3$\times$3, 6$\times$6, and 8$\times$8 features generated by spatial pyramid pooling~\cite{LazebnikSP06}.

Iterating $\ell$ from coarse to fine (from $1$ to $L$), the \emph{Context Aggregation module} refines $X^{(\ell)}$ using $\theta^{(\ell-1)}$ (while $X^{(1)}$ is refined by the initial context feature $\theta^{(0)}$);
the \emph{Context Updating module} then updates the context feature $\theta^{(\ell-1)}$ with the refined $X'^{(\ell)}$, forming the new context feature $\theta^{(\ell)}$ which facilitates finer-level semantic prediction by encapsulating the information from the coarsest to the current level.

Once the coarse-to-fine contextual module generates the refined feature pyramid $\{X'^{(\ell)}\}_{\ell=1,\ldots,L}$, the semantic pyramid $\{\hat{Y}^{(\ell)}\}_{\ell=1,\ldots,L}$ are predicted by
\begin{equation}
    \hat{Y}^{(\ell)} = \operatorname{ConvBlock}^{(\ell)}\left(X'^{(\ell)}\right) ,\label{eq:pred_sem}
\end{equation}
where $\operatorname{ConvBlock}^{(\ell)}$ consists of $\operatorname{Conv1x1}$, $\operatorname{BN}$, $\operatorname{ReLU}$, and a final $\operatorname{Conv1x1}$ projecting $D_s$ to the number of classes $C$.
In below, we detail the \emph{Context Aggregation module} ($\operatorname{CA}$) and \emph{Context Updating module} ($\operatorname{CU}$).

\paragraph{$\operatorname{CA}$---Context Aggregation module.}
The context aggregation module is illustrated in Fig.~\ref{fig:arch_ca}.
To refine $X^{(\ell)}$, we use attention operation to aggregate coarser-level context encoded in $\theta^{(\ell-1)}$.
Specifically, we transform $X^{(\ell)}$ to $X^{(\ell)}_{\text{query}}$,  $\theta^{(\ell-1)}$ to $\theta^{(\ell-1)}_{\text{key}}, \theta^{(\ell-1)}_{\text{value}}$ by $\operatorname{Conv1x1}$ layers; then we apply
\begin{align} 
    X_{\text{att}}^{(\ell)} &= \operatorname{Attention}\left(X_{\mathrm{query}}^{(\ell)}, \theta_{\mathrm{key}}^{(\ell-1)}, \theta_{\mathrm{value}}^{(\ell-1)}\right) , \\ 
    X'^{(\ell)} &= \operatorname{Conv1x1}_{\text{agg.}}^{(\ell)}\left(\operatorname{concat}\left(X_{\mathrm{att}}^{(\ell)}, X^{(\ell)}\right)\right) ,
\end{align}
where the $\operatorname{Attention}$ is the attention operation~\cite{VaswaniSPUJGKP17}, $\operatorname{Conv1x1}_{\text{agg.}}^{(\ell)}$ consisting of $\operatorname{Conv1x1}, \operatorname{BN}, \operatorname{ReLU}$ projects the concatenated $2D_s$ channels back to $D_s$.

\paragraph{$\operatorname{CU}$---Context Updating module.}
The context updating module is illustrated in Fig.~\ref{fig:arch_cu}.
To update the context feature $\theta^{(\ell-1)}$ on the refined feature $X'^{(\ell)}$ at level $\ell$, we apply
\begin{align}
    \theta^{(\ell)}_{\text{init}} &= \operatorname{PyramidPool}\left(X'^{(\ell)}\right) ,\\
    \theta^{(\ell)} &= \operatorname{Conv1x1}_{\text{upd.}}^{(\ell)}\left(\operatorname{concat}\left(\theta^{(\ell)}_{\text{init}}, \theta^{(\ell-1)}\right)\right) ,
\end{align}
where the $\operatorname{Conv1x1}_{\text{upd.}}^{(\ell)}$ consisting of $\operatorname{Conv1x1}, \operatorname{BN}, \operatorname{ReLU}$ projects the concatenated $2D_s$ channels back to $D_s$.
Note that the new context feature $\theta^{(\ell)}$ remains at the same low spatial resolution as $\theta^{(\ell-1)}$, so the overall coarse-to-fine contextual module runs efficiently.


\section{Experiments} \label{sec:exp}

We first introduce our implementation details in Sec.~\ref{ssec:imp_detail}.
Then, we report our comparison with state-of-the-art methods on three datasets in Sec.~\ref{ssec:exp_main} and the comparison on computation efficiency in Sec.~\ref{ssec:speed}.
Finally, thorough ablation study and performance analysis are conducted to support the contribution of our designed components in Sec.~\ref{ssec:exp_abla} and Sec.~\ref{ssec:exp_analy}, respectively.

\subsection{Implementation detail} \label{ssec:imp_detail}

\subsubsection{Training setting}
We mainly follow the training protocol of the public implementation of HRNet-OCR.
The SGD optimizer with momentum $0.9$ is employed.
Data augmentation includes random brightness, random left-right flip, random scaling with factor uniformly sampled from $[0.5, 2.0]$, and finally random crop to a fixed size.
The crop size, weight decay, and batch-size are set to $(512\times512, 1\mathrm{e}{-4}, 16)$ for all datasets.
The base learning rate and the number of epochs are set to $(0.02, 120)$, $(0.001, 110)$, and $(0.001, 200)$ for ADE20K, COCO-Stuff, and Pascal-Context, respectively.
The learning rate follows the poly schedule with the power factor set to $0.9$.

\subsubsection{Backbone setting}
We experiment with two backbone networks---HRNet48~\cite{WangSCJDZLMTWLX19} and ResNet101~\cite{HeZRS16}.
For simplicity, we ensure both backbones generate features at the same spatial level as the finest pyramid level, \ie, output stride $4$ in our experiments.

\paragraph{HRNet48.}
HRNet~\cite{WangSCJDZLMTWLX19} provides high-resolution features of output stride $4$, so we directly attach our \emph{Unity head} and \emph{Semantic head} to the end of HRNet.

\paragraph{ResNet101.}
ResNet~\cite{HeZRS16} produces coarse features of output stride $32$.
To obtain better results, some recent methods~\cite{LiZHTTY20,YuanCW20,LiuHQRL20,HuJGBWY20,LiLZCSLTT20} employ the dilated version of ResNet with output stride $8$.
However, we find such modification leads to lower speed and more memory footprint, so we adopt the standard ResNet with a lightweight decoder.
We simply reduce the number of channels of each stage from ResNet to save computation and apply the fusion module~\cite{WangSCJDZLMTWLX19} as the decoder to form feature of output stride $4$.
Comparing to the dilated ResNet, our adaptation requires only $0.75\times$ processing time and $0.88\times$ memory footprint (see the supplementary material for details about implementation and computational efficiency).
Besides, some recent methods~\cite{HuJGBWY20,LiLZCSLTT20} also adopt ASPP~\cite{ChenZPSA18}, which we do not employ, for the ResNet backbone.

\begin{table}
  \centering
  \begin{tabular}{l@{\hskip 3pt}ll|c}
    \hline
    Method & Venue & Backbone & mIoU (\%) \\
    \hline
    CFNet~\cite{ZhangZWX19}              & CVPR2019    & ResNet101 & 44.89 \\
    APCNet~\cite{HeDZW019}               & CVPR2019    & ResNet101 & 45.38 \\
    CCNet~\cite{HuangWHHW019}            & ICCV2019    & ResNet101 & 45.22 \\
    ANL~\cite{ZhuXBHB19}                 & ICCV2019    & ResNet101 & 45.24 \\
    ACNet~\cite{FuLW0BTL19}              & ICCV2019    & ResNet101 & 45.90 \\
    CPNet~\cite{YuWGYSS20}               & CVPR2020    & ResNet101 & \underline{46.27}\\
    SPNet~\cite{HouZCF20}                & CVPR2020    & ResNet101 & 45.60\\
    QGN~\cite{ChittaAH20}                & WACV2020    & ResNet101 & 43.91\\
    GFFNet~\cite{LiZHTTY20}              & AAAI2020    & ResNet101 & 45.33\\
    OCR~\cite{YuanCW20}                  & ECCV2020    & ResNet101 & 45.28\\
    DNL~\cite{YinYCLZLH20}               & ECCV2020    & ResNet101 & 45.97\\
    CaCNet~\cite{LiuHQRL20}              & ECCV2020    & ResNet101 & 46.12\\
    ours                                 & -           & ResNet101 & \textbf{47.00}\\
    \hline
    HRNet~\cite{WangSCJDZLMTWLX19}       & TPAMI2019   & HRNet48   & 44.20\\
    CCNet~\cite{HuangWHHW019}\textdagger & -           & HRNet48   & 45.65\\
    ANL~\cite{ZhuXBHB19}\textdagger      & -           & HRNet48   & 45.23\\
    OCR~\cite{YuanCW20}                  & ECCV2020    & HRNet48   & 45.50\\
    DNL~\cite{YinYCLZLH20}               & ECCV2020    & HRNet48   & \underline{45.82}\\
    ours                        & -           & HRNet48   & \textbf{47.16}\\
    \hline
    \multicolumn{4}{l}{\textdagger\footnotesize{Our reproduction by replacing the backbone with HRNet48}}
  \end{tabular}
  \caption{
    Comparisons on ADE20K~\cite{ZhouZPFB017} validation set.
  }
  \label{table:exp_ade20k}
  \vspace{-1.5em}
\end{table}

\begin{table}
  \centering
  \begin{tabular}{l@{\hskip 3pt}ll|c}
    \hline
    Method & Venue & Backbone & Score \\
    \hline
    PSPNet~\cite{ZhaoSQWJ17} & CVPR2017 & ResNet269 & 55.38 \\
    EncNet~\cite{ZhangDSZWTA18} & CVPR2018 & ResNet101 & 55.67 \\
    ACNet~\cite{FuLW0BTL19} & ICCV2019 & ResNet101 & 55.84 \\
    DNL~\cite{YinYCLZLH20} & ECCV2020 & ResNet101 & 56.23 \\
    ours & - & ResNet101 & {\bf 56.67} \\
    \hline
    DNL~\cite{YinYCLZLH20} & ECCV2020 & HRNet48 & 55.98 \\
    ours & - & HRNet48 & {\bf 58.04} \\
    \hline
  \end{tabular}
  \caption{ADE20K~\cite{ZhouZPFB017} official evaluation.}
  \label{table:ade20k_official}
  \vspace{-2.5em}
\end{table}

\vspace{-1em}

\subsubsection{\emph{Specialize and Fuse} setting}
The backbone features have $D=720$ channels for both our HRNet48 and ResNet101 experiments,
and we set $D_u$ in the unity head to $64$ and $D_s$ in the semantic pyramid head to $512$.
We use the instantiation of $L=4$ with output strides $\{4,8,16,32\}$ for the semantic pyramid.
We set a high threshold $\tau=0.9$ for the binary classifier in the unity pyramid to suppress false positives for the unity-cell prediction as, intuitively, a false positive always introduces error while a false negative could have the chance to be ``remedied'' by finer level semantic.


\begin{table}
  \centering
  \begin{tabular}{l@{\hskip 3pt}ll|c}
    \hline
    Method & Venue & Backbone & mIoU (\%) \\
    \hline
    SVCNet~\cite{DingJSLW19}             & CVPR2019    & ResNet101 & 39.6\\
    DANet~\cite{FuLT0BFL19}              & CVPR2019    & ResNet101 & 39.7\\
    EMANet~\cite{LiZWYLL19}              & ICCV2019    & ResNet101 & 39.9\\
    ACNet~\cite{FuLW0BTL19}              & ICCV2019    & ResNet101 & 40.1\\
    GFFNet~\cite{LiZHTTY20}              & AAAI2020    & ResNet101 & 39.2\\
    OCR~\cite{YuanCW20}                  & ECCV2020    & ResNet101 & 39.5\\
    CDGCNet~\cite{HuJGBWY20}             & ECCV2020    & ResNet101 & \textbf{40.7}\\
    ours                                 & -           & ResNet101 & \textbf{40.7}\\
    \hline
    HRNet~\cite{WangSCJDZLMTWLX19}       & TPAMI2019   & HRNet48   & 37.9\\
    CCNet~\cite{HuangWHHW019}\textdagger & -           & HRNet48   & 39.8\\
    ANL~\cite{ZhuXBHB19}\textdagger      & -           & HRNet48   & \underline{40.6}\\
    OCR~\cite{YuanCW20}                  & ECCV2020    & HRNet48   & \underline{40.6}\\
    ours                        & -           & HRNet48   & \textbf{41.0}\\
    \hline
    \multicolumn{4}{l}{\textdagger\footnotesize{Our reproduction by replacing the backbone with HRNet48}}
  \end{tabular}
  \caption{
    Comparisons on COCO-Stuff~\cite{CaesarUF18} test set.
  }
  \label{table:exp_cocostuff10k}
  \vspace{-1em}
\end{table}

\begin{table}
  \centering
  \begin{tabular}{l@{\hskip 3pt}ll|c}
    \hline
    Method & Venue & Backbone & mIoU (\%) \\
    \hline
    CFNet~\cite{ZhangZWX19}              & CVPR2019    & ResNet101 & 54.0\\
    APCNet~\cite{HeDZW019}               & CVPR2019    & ResNet101 & 54.7\\
    SVCNet~\cite{DingJSLW19}             & CVPR2019    & ResNet101 & 53.2\\
    DANet~\cite{FuLT0BFL19}              & CVPR2019    & ResNet101 & 52.6\\
    BFP~\cite{DingJLM019}                & ICCV2019    & ResNet101 & 53.6\\
    ANL~\cite{ZhuXBHB19}                 & ICCV2019    & ResNet101 & 52.8\\
    EMANet~\cite{LiZWYLL19}              & ICCV2019    & ResNet101 & 53.1\\
    ACNet~\cite{FuLW0BTL19}              & ICCV2019    & ResNet101 & 54.1\\
    DGCNet~\cite{ZhangLAYTT19}           & BMVC2019    & ResNet101 & 53.7\\
    CPNet~\cite{YuWGYSS20}               & CVPR2020    & ResNet101 & 53.9\\
    SPNet~\cite{HouZCF20}                & CVPR2020    & ResNet101 & 54.5\\
    GFFNet~\cite{LiZHTTY20}              & AAAI2020    & ResNet101 & 54.2\\
    OCR~\cite{YuanCW20}                  & ECCV2020    & ResNet101 & 54.8\\
    DNL~\cite{YinYCLZLH20}               & ECCV2020    & ResNet101 & 54.8\\
    CaCNet~\cite{LiuHQRL20}              & ECCV2020    & ResNet101 & \underline{55.4}\\
    ours                                 & -           & ResNet101 & \textbf{55.6}\\
    \hline
    HRNet~\cite{WangSCJDZLMTWLX19}       & TPAMI2019   & HRNet48   & 54.0\\
    OCR~\cite{YuanCW20}                  & ECCV2020    & HRNet48   & \underline{56.2}\\
    DNL~\cite{YinYCLZLH20}               & ECCV2020    & HRNet48   & 55.3\\
    ours                        & -           & HRNet48   & \textbf{57.0}\\
    \hline
  \end{tabular}
  \caption{
    Comparisons on Pascal-Context~\cite{MottaghiCLCLFUY14} test set.
  }
  \label{table:exp_pascalctx}
  \vspace{-1.5em}
\end{table}

\subsection{Comparison with state-of-the-arts} \label{ssec:exp_main}
Following the literature, we apply multi-scale and left-right flip testing augmentation to report our results.

\paragraph{ADE20K~\cite{ZhouZPFB017}.}
ADE20k is a dataset with diverse scenes containing 35 stuff and 115 thing classes. The training, validation, and test split contains 20K/2K/3K images, respectively.
The results on validation set of ADE20K is shown in Table~\ref{table:exp_ade20k}.
Our method establishes new state-of-the-art results with both ResNet101 and HRNet48 backbone.
In addition, we submit our prediction on hold-out test set to ADE20k's server for official evaluation.
The results reported in Table~\ref{table:ade20k_official} also show our advantage over previous methods.

\paragraph{COCO-Stuff~\cite{CaesarUF18}.}
COCO-Stuff is a challenging dataset with 91 stuff and 80 thing classes. The training and the test sets contain 9K and 1K images, respectively.
In Table~\ref{table:exp_cocostuff10k}, our approach shows comparable performance to recent state-of-the-art results with ResNet101 backbone and outperforms previous methods with HRNet48 backbone.

\paragraph{Pascal-Context~\cite{MottaghiCLCLFUY14}.}
Pascal-Context is a widely-used dataset for semantic segmentation. It contains 59 classes and one background class, consisting of 4{,}998 training and 5{,}105 test images.
Results on Pascal-Context test set is presented in Table~\ref{table:exp_pascalctx}.
With both ResNet101 and HRNet48 backbone, our method achieves state-of-the-art results comparing to previous methods with the same backbone.

\subsection{Computational efficiency} \label{ssec:speed}
We compare the testing FPS, training iterations per second, and GPU memory consumption by our in-house implementation in Table~\ref{table:speed_mem}.
Our full approach shows similar computational efficiency to the recent efficient variant of self-attention modules~\cite{HuangWHHW019,ZhuXBHB19,YuanCW20} for semantic segmentation, meanwhile showing better accuracy (Table~\ref{table:exp_ade20k} and Table~\ref{table:exp_cocostuff10k}).
\begin{table}[h]
    \centering
    \begin{tabular}{l|c|cc}
    \hline
    \multirow{2}{*}{Method} & Testing & \multicolumn{2}{c}{Training} \\
    & FPS$\uparrow$ & it./sec.$\uparrow$ & Mem.$\downarrow$ \\
    \hline
    HRNet48~\cite{WangSCJDZLMTWLX19}    & 28 & 2.6 & 8.2G \\
    \hline
    HRNet48 + CCNet~\cite{HuangWHHW019} & 21 & 1.7 & 9.9G \\
    HRNet48 + ANL~\cite{ZhuXBHB19}      & 26 & 2.2 & 8.6G \\
    HRNet48 + OCR~\cite{YuanCW20}       & 24 & 2.1 & 9.6G \\
    HRNet48 + ours                      & 24 & 2.0 & 8.9G \\
    \hline
    \end{tabular}
    \caption{
        Comparing the model efficiency measured on a GeForce RTX 2080 Ti with image size $512 \times 512$.
        Testing FPS is averaged for processing $50$ images.
        Model training is monitored with a batch size of $4$.
    }
    \label{table:speed_mem}
    \vspace{-1em}
\end{table}


\vspace{-1em}
\subsection{Ablation study} \label{ssec:exp_abla}
We conduct comprehensive ablation experiments to verify the effectiveness of our proposals.
In our ablation experiments, the HRNet32 backbone is employed, and we sub-sample ADE20K's original training split into 16K/4K for training and validation.
We put detailed description and architecture diagram for each experiment in the supplementary material and focus on the comparisons and discussion here.

\paragraph{The effectiveness of the pyramidal output representation.}
As demonstrated in Table~\ref{table:abla_pyramid_contex}, our pyramidal ``output'' representation (the second row) consistently brings improvement over the standard single-level output (the first row) under various model settings.
It gains $+1.65$, $+1.05$ and $+2.31$ mIoU improvement without any contextual module, with ANL module~\cite{ZhuXBHB19} and with the proposed coarse-to-fine contextual module, respectively.


\paragraph{The effectiveness of the contextual module.}
Our contextual module is designed to accord with the multi-level essence of the proposed pyramidal output.
As shown in Table~\ref{table:abla_pyramid_contex}, when predicting the standard output (of a single finest level), our coarse-to-fine contextual module and the ANL~\cite{ZhuXBHB19} gain similar improvement over the baseline: $+1.58$ ($40.42\rightarrow 42.00$) and $+1.60$ ($40.42\rightarrow 42.02$).
However, when predicting the proposed pyramidal output, our coarse-to-fine contextual module achieves a more significant $+2.24$ improvement ($42.07\rightarrow 44.31$) than appending a single ANL to the backbone $+1.00$ ($42.07\rightarrow 43.07$).
Furthermore, our module also gains more improvement than applying ANL modules at all pyramid level (denoted as ANL-multi) $+1.38$ ($42.07\rightarrow 43.45$), indicating the effectiveness of our design to aggregate contextual information from coarser pyramid levels.

\begin{table}[h]
    \vspace{-0.5em}
    \centering
    \begin{tabular}{c|c@{\hskip 6pt}c@{\hskip 6pt}c@{\hskip 6pt}c}
    \hline
    & \multicolumn{4}{c}{Contextual module} \\
    Output format & - & ANL & ANL-multi & ours \\
    \hline
    Single (standard) & 40.42 & 42.02 & - & 42.00 \\
    Pyramidal (ours)  & 42.07 & 43.07 & 43.45 & 44.31 \\
    \hline
    \end{tabular}
    \caption{
        Ablation study for the main proposals in two aspects---{\it i)} the pyramidal output representation (the rows) and {\it ii)} the contextual module (the columns).
        Mean IoU (\%) of each setting is reported for comparison.
        Here ``ANL'' is appending a single ANL to the backbone, and ``ANL-multi'' is applying ANL to each level of our pyramidal ``output''.
    }
    \label{table:abla_pyramid_contex}
    \vspace{-2em}
\end{table}

\paragraph{Supervision for specialization of different pyramid output levels.}
To enforce specialization in training phase, it is important to relabel cells which are ``done by coarser'' as ``don't care''.
In Table~\ref{table:abla_train}, we show the results comparing to other relabeling policies.
We can see that naively training all semantic pyramid levels to predict full semantic segmentation map do not achieve any improvement.
This is reasonable as it ignores the fact that only a small portion of the cells in finer pyramid level would be activated in testing.
A simple fix is explicitly relabeling all descendants of a ground-truth unity-cell as ``don't care'' during training.
By doing so, different semantic pyramid levels now can specialize in the pixels assigned by the oracle.
A clear $+1.12$ mIoU improvement is now shown.
However, the simple fix still ignores the fact that the unity-cell prediction may produce false negatives where a ground-truth unity-cell is falsely classified as a mix-cell and thus a finer-level semantic prediction is referred in inference phase.
As a result, the false negatives might make the train-test distributions inconsistent.
Finally, our design of re-labeling the descendants of a true positive unity-cell as ``don't care'' gains an extra $+0.53$ mIoU improvement.

\begin{table}[h]
    \vspace{-0.5em}
  \centering
  \begin{tabular}{c|ccc}
    \hline
    single output & naive & simple fix & our final\\
    \hline
    40.42 & 40.41 & 41.54 & 42.07 \\
    \hline
  \end{tabular}
  \caption{
    Comparing the relabeling policy in the training procedures for our pyramidal output (the right-most 3 columns).
    No contextual module is deployed for all experiments.
  }
  \label{table:abla_train}
  \vspace{-2em}
\end{table}


\paragraph{Does the improvement stem from auxiliary supervision?}
One may argue that the improvement from our pyramidal output representation stems from the rich supervision at the multi-level rather than our ``specialize and fuse" process.
To clarify the contribution, 
we skip the ``fuse" phase at inference and use the semantic output at the finest-level as the final output. In this case, the supervision at multi-level are considered as auxiliary supervision.
In Table~\ref{table:abla_aux}, we can see that the auxiliary supervision indeed improves the performance ($+0.63$). However, the improvement is not comparable to our {\it specialized and fuse} process ($+2.31$), which suggests that the main contributor to our superior performance is not multi-level supervision but the {\it specialized and fuse} strategy.

\begin{table}[h]
    \vspace{-0.5em}
    \centering
    \begin{tabular}{c|cc}
    \hline
    single output & aux. supervision & specialize \& fuse \\
    \hline
    42.00 & 42.63 (+0.63) & 44.31 (+2.31) \\ 
    \hline
    \end{tabular}
    \caption{
        How to use coarser-level outputs.
        The proposed contextual module is deployed for all results.
    }
    \label{table:abla_aux}
    \vspace{-2em}
\end{table}


\paragraph{Number of pyramid levels in our output representation.}
This work focuses on the setting of $L{=}4$ pyramid levels of output strides $\{4,8,16,32\}$ for our pyramidal output representation in all experiments.
We also experiment with $L{=}2$ levels of output strides $\{4,32\}$ which yields a slightly worse results ($44.31$ vs. $44.20$ mIoU).
Therefore, we stick to the $L{=}4$ setting.

\subsection{Performance analysis} \label{ssec:exp_analy}

\paragraph{Does each pyramid level specializes in the selected pixels by unity pyramid?}
In Table~\ref{table:miou_mat}, we divide pixels into four groups (the four columns, $\ell \in \{1,2,3,4\}$) according to the assignment from our predicted unity pyramid (Sec.~\ref{ssec:fuse}).
For each group, we show the performance of the semantic predictions of the four pyramid levels (the four rows, $\ell' \in \{1,2,3,4\}$).
We can see that the mIoU is degraded if a pixel refers to a pyramid level that does not agree with the assignment by the trained unity pyramid ($\ell' \neq \ell$ in each column of Table.~\ref{table:miou_mat}), indicating that different semantic pyramid levels learn to specialize in predicting the pixels assigned by our predicted unity pyramid.

\begin{table}[h]
  \vspace{-0.5em}
  \centering
  \begin{tabular}{c|c|c|c|c}
        \hline
        $\ell'$~\textbackslash~$\ell$ & 4 & 3 & 2 & 1 \\
        \hline
        4 & \textbf{33.48} & 39.52 & 42.13 & 38.61 \\
        3 & 31.63 & \textbf{41.17} & 45.98 & 44.57 \\
        2 & 27.31 & 38.13 & \textbf{46.34} & 47.52 \\
        1 & 21.94 & 28.94 & 40.05 & \textbf{48.05} \\
        \hline
    \end{tabular}
    \caption{
        mIoU over a pair of pyramid levels, where $\ell$ indicates the level of the unity-cell prediction and $\ell^{'}$ indicates the level of the semantic prediction. When $\ell^{'} = \ell$, the mIoU is higher than other $\ell^{'} \neq \ell$ for every $\ell$.
    }
    \label{table:miou_mat}
    \vspace{-2em}
\end{table}

\begin{figure}[h]
  \vspace{-.5em}
  \centering
  \includegraphics[width=.9\linewidth]{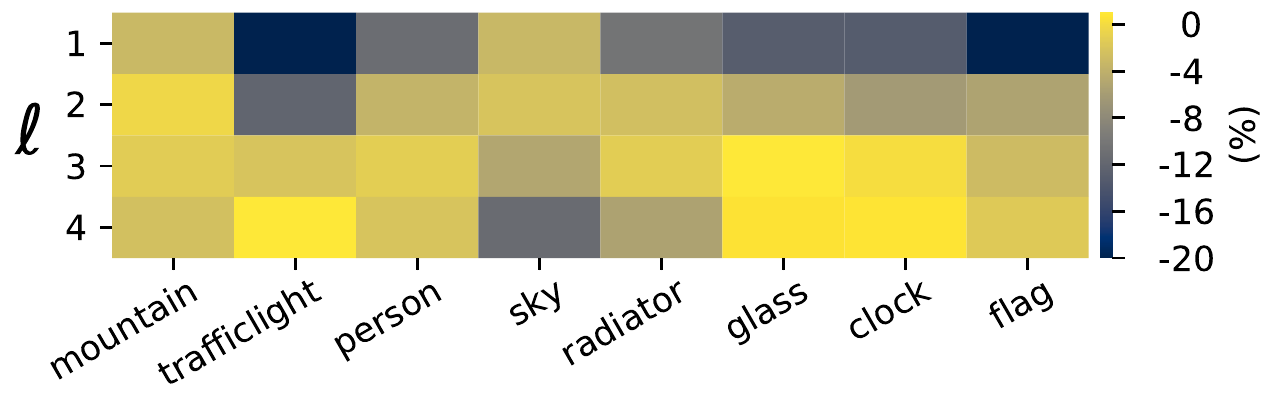}
  \vspace{-1em}
  \caption{
  Semantic segmentation performance at each level $\{\hat{Y}^{(\ell)}\}_{\ell=1,\ldots,4}$ on different classes.
  We show the IoU degradation due to using a level $\hat{Y}^{(\ell)}$ versus using the fused $\hat{Y}$.
  }
  \label{fig:scales_class}
\end{figure}

\paragraph{Do different pyramid levels specialize in different classes?}
To demonstrate our intuition that different pyramid levels have their specializations in different classes, in each row of Fig.~\ref{fig:scales_class}, we show the per-class IoUs predicted by each semantic level.
The results suggest that each $\hat{Y}^{(\ell)}$ learns better for different classes in practice.
For instance, $\hat{Y}^{(4)}$ performs better at {\sf trafficlight}, while $\hat{Y}^{(2)}$ is good at {\sf mountain}.
See supplementary material for more visualizations.

\paragraph{Qualitative results.}
We show some qualitative results comparing to the strong baseline, HRNet48-ANL, in Fig.~\ref{fig:qualitative}.
See the supplementary material for more examples.

\begin{figure}[h]
   \scriptsize
  \vspace{-1em}
   \centering
    \begin{tabular}{c@{\hskip 1pt}c@{\hskip 1pt}c@{\hskip 1pt}c}
        input & ground truth & HRNet48-ANL & HRNet48-ours \\
        \includegraphics[width=0.23\linewidth]{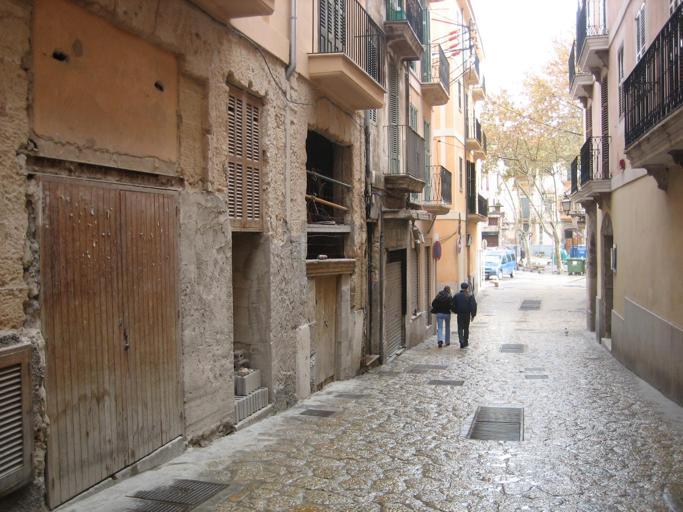}&
        \includegraphics[width=0.23\linewidth]{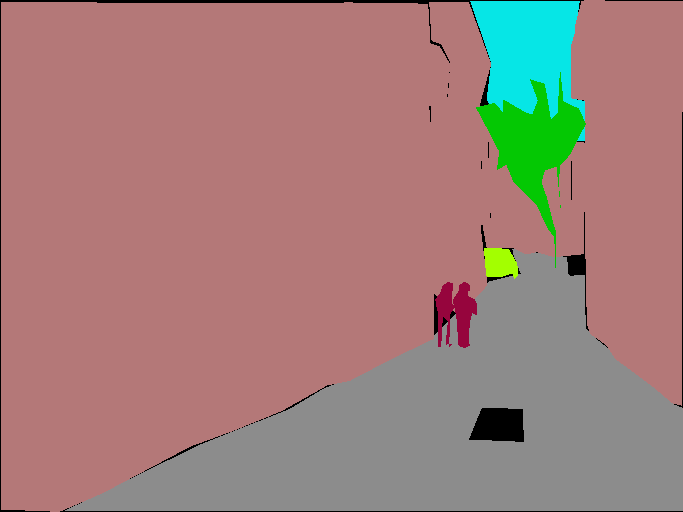}&
        \includegraphics[width=0.23\linewidth]{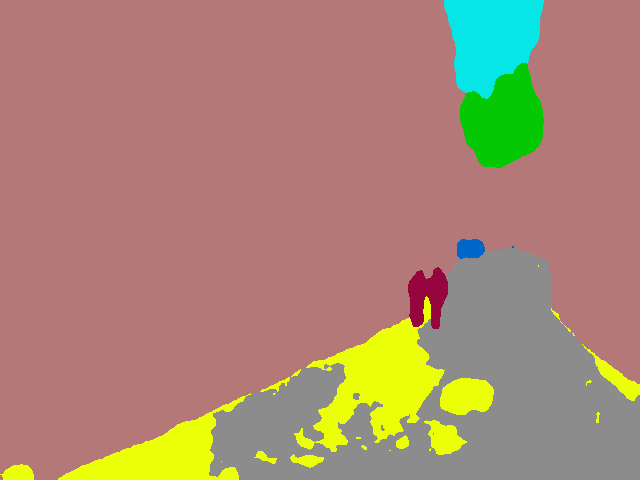}&
        \includegraphics[width=0.23\linewidth]{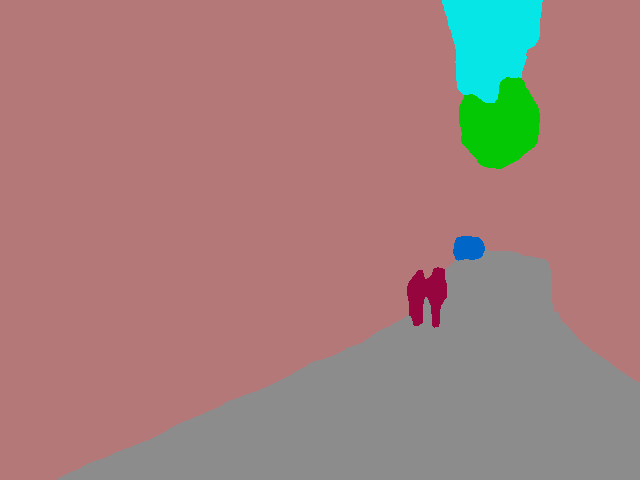}\\
        \includegraphics[width=0.23\linewidth]{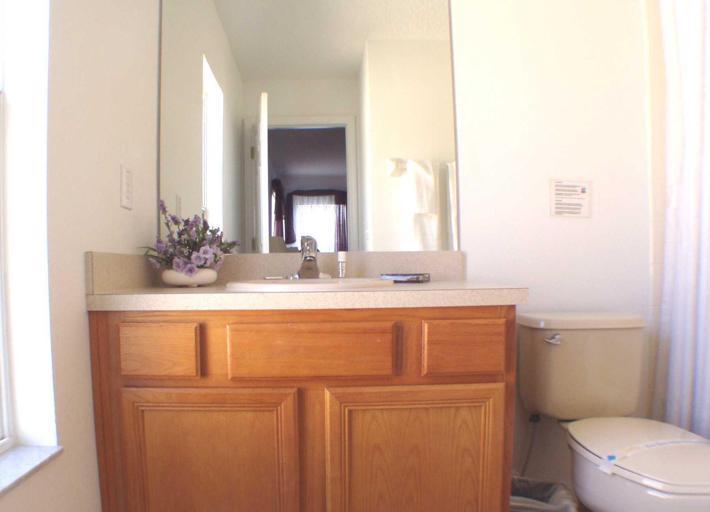}&
        \includegraphics[width=0.23\linewidth]{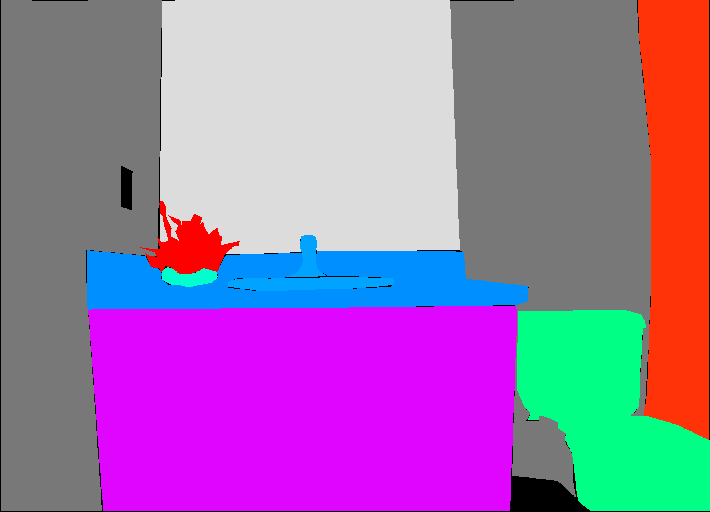}&
        \includegraphics[width=0.23\linewidth]{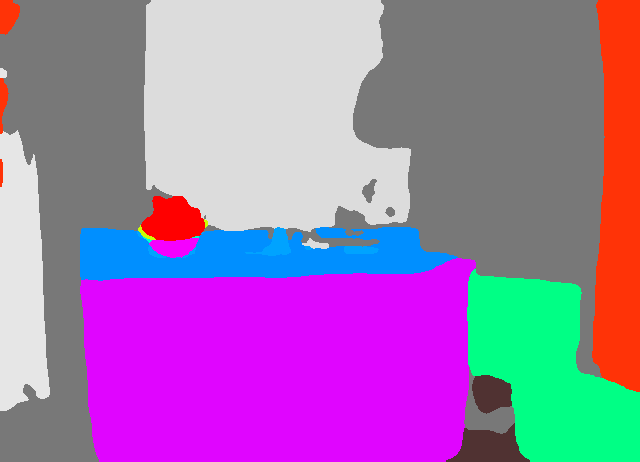}&
        \includegraphics[width=0.23\linewidth]{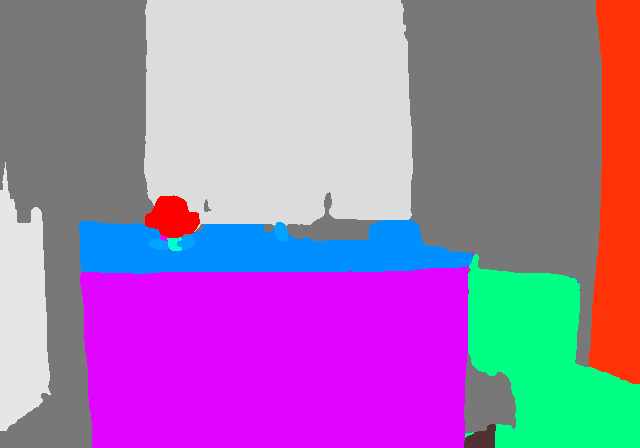}\\
        \includegraphics[width=0.23\linewidth]{figs/qualitative/qual/ADE_val_00000085.jpg}&
        \includegraphics[width=0.23\linewidth]{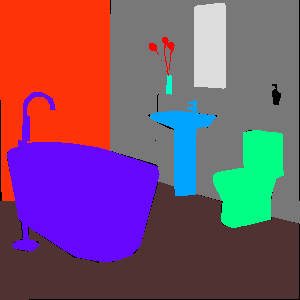}&
        \includegraphics[width=0.23\linewidth]{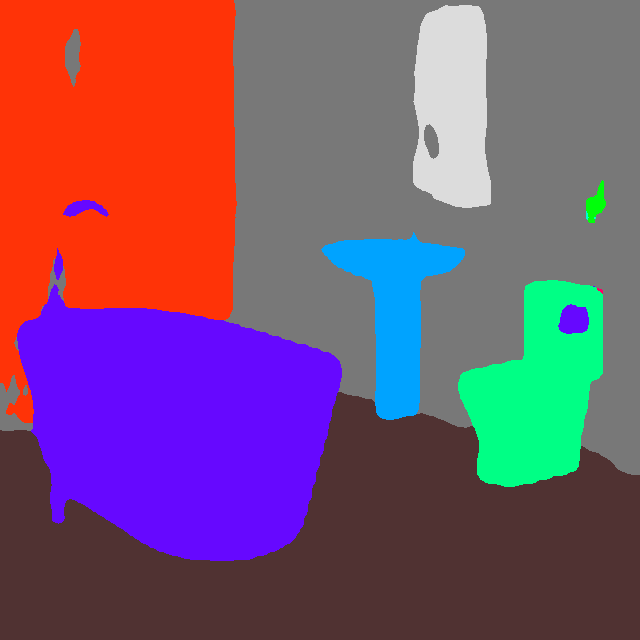}&
        \includegraphics[width=0.23\linewidth]{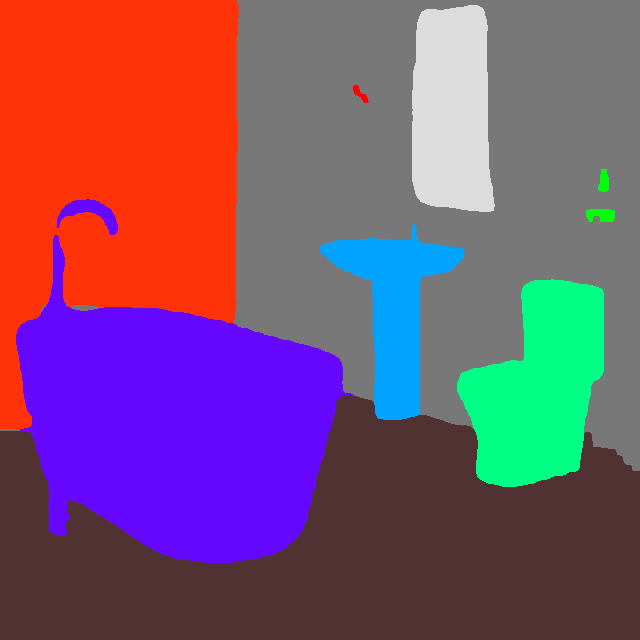}\\
    \end{tabular}
    \vspace{-1.5em}
    \caption{
        Qualitative results.
        In the examples, our approach yields better results in the stuff (the $1^{\text{st}}$ row), the large thing (the $2^{\text{nd}}$ row), and the thin part of the object (the $3^{\text{rd}}$ row).
    }
    \vspace{-1em}
    \label{fig:qualitative}
\end{figure}



\section{Conclusion} \label{sec:conclusion}
We present a novel ``output" representation for the task of semantic segmentation. The proposed pyramidal output format and the fusing procedure follow the motivation to assign each pixel to an appropriate pyramid level for better specialization and the parsimony principle. We also present a contextual module, which is efficient and fits the essence of the proposed pyramidal output, improving our performance further. Improvements are shown through extensive experiments.
Finally, our performance is on par with or better than the recent state-of-the-art on three widely-used semantic segmentation datasets.

\vspace{1em}
\noindent \textbf{Acknowledgements:}
This work was supported in part by the MOST, Taiwan under Grants 110-2634-F-001-009 and 110-2634-F-007-016, MOST Joint Research Center for AI Technology and All Vista Healthcare. We thank National Center for High-performance Computing (NCHC) for providing computational and storage resources.

{\small
\bibliographystyle{ieee_fullname}
\bibliography{egbib}
}

\clearpage
\setcounter{section}{0}
\setcounter{table}{0}
\setcounter{figure}{0}
\renewcommand\thesection{\Alph{section}}
\renewcommand\thetable{\Alph{table}}
\renewcommand\thefigure{\Alph{figure}}

\section{Visualization of the intermediate results}
\begin{figure}[h]
  \centering
  \vspace{-1em}
  \includegraphics[width=0.87\linewidth]{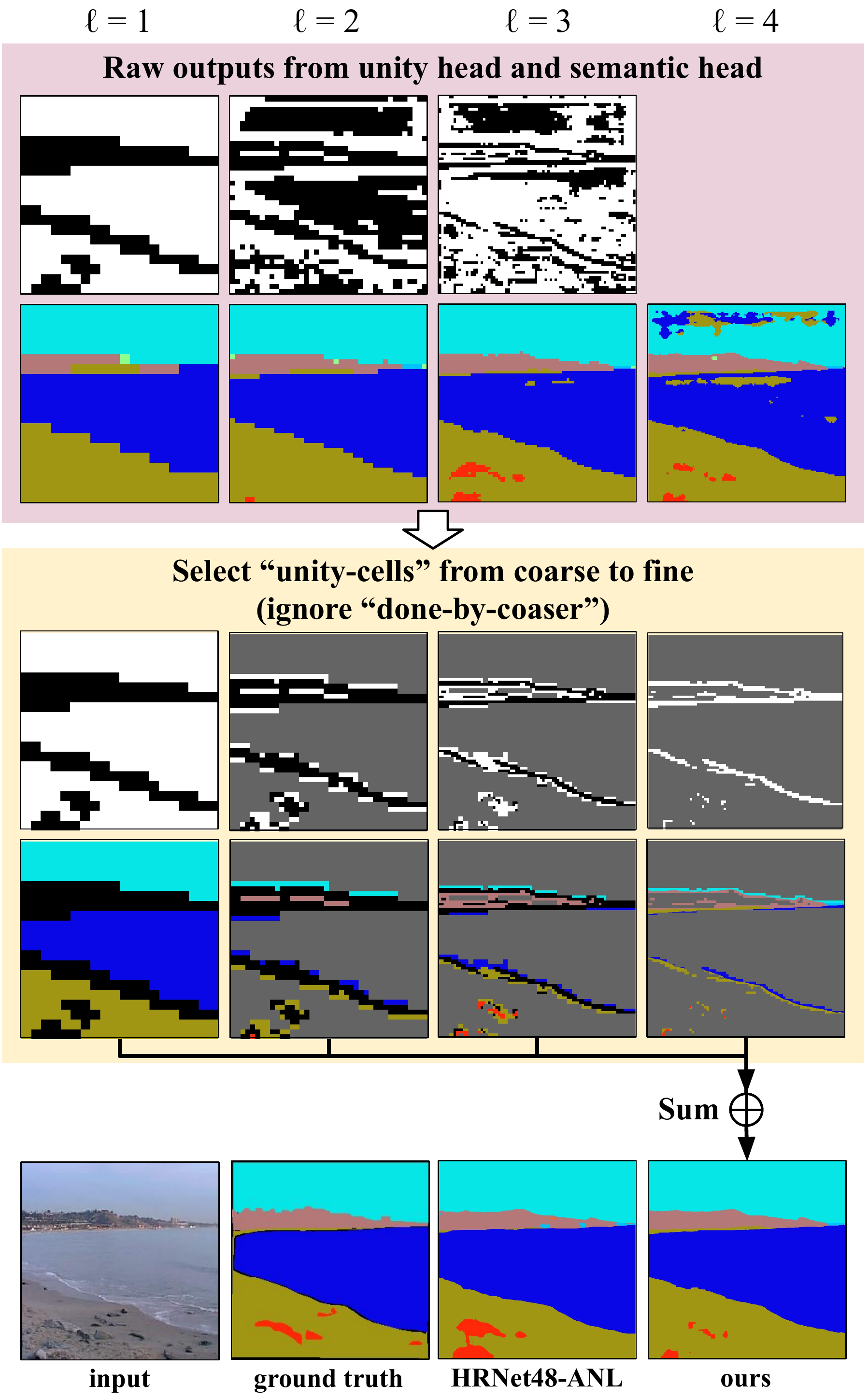}
  \caption{
    An example from the ADE20K validation set showing the intermediate results.
    \textbf{Top}: The raw semantic pyramid outputs and raw unity pyramid outputs.
    \textbf{Middle}: From the coarsest output to the finest output, we select semantic labels only from ``unity-cells'' but ignore the ``done-by-coarser'' regions (colored in grey).
    \textbf{Bottom}: Input image, ground truth, HRNet48-ANL prediction, and our final fused prediction.
  }
  \label{fig:inter_112}
\end{figure}

We show the intermediate results of our approach and illustrate how they are fused in Fig.~\ref{fig:inter_112}.

\section{Pyramidal ground truth}
Here, we detail how to generate the pyramidal ground truth from standard per-pixel semantic labels when there are ``don't care'' annotations in the original dataset.
For a cell covering $s_{\ell} \times s_{\ell}$ pixels, we consider 3 cases for different treatments.
\begin{itemize}
    \item \textbf{All pixels share a semantic class.} In this case, the ground-truth label for the unity pyramid is ``unity-cell'' (\ie, positive), and the label for the semantic pyramid is the shared class.
    \item \textbf{More than one semantic classes appear.} In this case, the label for the unity pyramid is ``mix-cell'' (\ie, negative), and the label for the semantic pyramid is ``don't care'' because it is ambiguous to use one semantic label to represent all underlying pixels of different semantic classes and thus a finer semantic prediction should be referred to.
    \item \textbf{One semantic class and ``don't care'' appear.} Both the unity and the semantic ground truth are defined as ``don't care'' as it is ambiguous to determine whether the cell is ``unity-cell'' or ``mix-cell''. During the training re-labeling procedure, such a cell is never regarded as true positive, and thus its children cells in the next finer level would never be re-labeled as ``done by coarser''.
\end{itemize}


\begin{table*}
  \centering
  \begin{tabular}{lcccc}
    \hline
    ID & mIoU (\%) & Contextual module & Output format & Training procedure\\
    \hline
    \textit{A} & 40.42  & - & single (standard) & - \\
    \textit{B} & 42.02 & ANL & single (standard) & - \\
    \textit{C} & 42.00 & ours & single (standard) & - \\
    \hline
    \textit{D} & 40.41 & - & pyramidal (ours) \{4,8,16,32\} & naive \\
    \textit{E} & 41.54 & - & pyramidal (ours) \{4,8,16,32\} & simple fix \\
    \hline
    \textit{F} & 42.07 & - & pyramidal (ours) \{4,8,16,32\} & our final \\
    \textit{G} & 43.07 & ANL & pyramidal (ours) \{4,8,16,32\} & our final \\
    \textit{H} & 43.45 & ANL-multi & pyramidal (ours) \{4,8,16,32\} & our final \\
    \textit{I} & \textbf{44.31} & ours & pyramidal (ours) \{4,8,16,32\} & our final \\
    \hline
    \textit{J} & 42.63 & ours & single (standard) & \{4,8,16\} only for aux. \\
    \hline
    \textit{K} & 44.20 & ours & pyramidal (ours) \{4,32\} & our final \\
    \hline
  \end{tabular}
  \caption{The ablation experiment results reorganized in a unified view.}
  \label{table:abla_main}
\end{table*}

\begin{figure*}
    \centering
    \includegraphics[width=.8\linewidth]{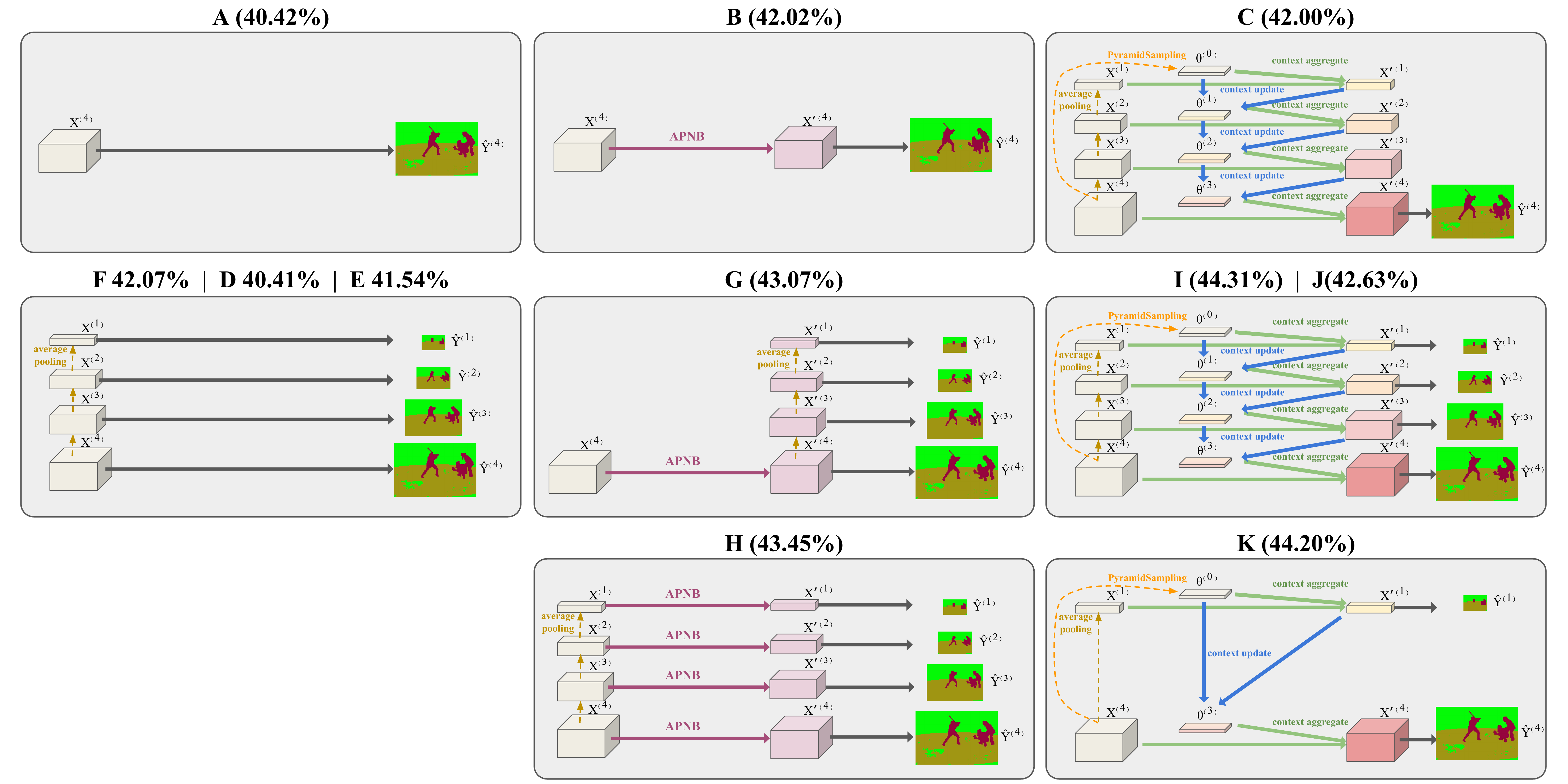}
    \caption{
        The network architectures of the ablation experiments listed in Table~\ref{table:abla_main}.
    }
    \label{fig:detail_abla}
\end{figure*}

\section{Detailed model settings in ablation study}
We reorganize the ablation experiment results presented in the main paper into a unified view in Table~\ref{table:abla_main} and illustrate their network architectures in Fig.~\ref{fig:detail_abla}.
We label each experiment with an ID  (A$\sim$K) and describe the detailed architecture setting below.
We call the layers projecting latent dimension $D_s$ to number of classes $C$ as \textit{final projection layer}, which is implemented as Conv1$\times$1 $\rightarrow$ BN $\rightarrow$ ReLU $\rightarrow$ Conv1$\times$1.

\begin{itemize}
    \item \textit{A} (40.42 \%) directly appends the \textit{final projection layer} to $X^{(4)}$.
    \item \textit{B} (42.02 \%) refines $X^{(4)}$ with ANL's APNB block.
    \item \textit{C} (42.00 \%) is our coarse-to-fine contextual module, but only a single finest level semantic prediction is outputted.
    \item \textit{F} (42.07 \%) performs average-pooling on $X^{(4)}$ to get features $X^{(1)}, X^{(2)}, X^{(3)}$ of desired spatial scales, and each of $X^{(1)},\ldots,X^{(4)}$ has its own \textit{final projection layer}.
    \item \textit{D} (40.41 \%): is a variant of \textit{F} where the training ground-truth is the raw pyramidal ground truth $Y^{(1)},\ldots,Y^{(4)}$ and $U^{(1)},\ldots,U^{(3)}$ without the re-labeling procedure to encourage specialization in each pyramid level.
    \item \textit{E} (41.54 \%) is a variant of \textit{F} where the ground-truth unity-cell, instead of the true positive unity-cell, is used for ``don't care'' re-labeling.
    \item \textit{G} (43.07 \%) refines $X^{(4)}$ of \textit{F} by appending ANL's APNB block to $X^{(4)}$.
    \item \textit{H} (43.45 \%) is similar to \textit{G} but appends APNB block to each of $X'^{(1)},X'^{(2)},X'^{(3)},X'^{(4)}$.
    \item \textit{I} (44.31 \%) is the final version of the proposed \textbf{Specialize and Fuse}.
    \item \textit{J} (42.63 \%) is similar to \textit{I}, but the non-finest level predictions $\hat{Y}^{(1)},\ldots,\hat{Y}^{(3)}$ are only used for auxiliary loss (the loss weight is set to $0.4$) in the training phase and discarded in the inference phase.
    \item \textit{K} (44.20 \%) is similar to \textit{I} but with less pyramid levels.
\end{itemize}

\clearpage
\section{Architecture of our ResNet-decoder}
The standard ResNet produces coarse features of output stride $32$.
To obtain better results, recent state-of-the-art methods employ the dilated version of ResNet, which generates features of output stride 8.
However, we find such a modification leads to a lower speed and more memory footprint as shown in Table~\ref{table:speed_mem} (the $2^{\text{nd}}$ row), so we adopt the standard ResNet with a lightweight decoder instead.
Specifically, we use the features from the four stages of a standard ResNet, which respectively have output strides $4, 8, 16, 32$, and $256, 512, 1024, 2048$ latent channels.
To reduce the computational cost, the numbers of latent channels are decreased from $[256, 512, 1024, 2048]$ to $[48, 96, 192, 384]$ with $\operatorname{Conv3x3}$ layers.
We then fuse the four features by applying the last stage of HRNet, which produces a high-resolution feature for the following semantic segmentation model head.
The comparison of computational efficiency between ResNet variants is shown in Table~\ref{table:speed_mem}.
Note that the constructed ResNet-decoder runs faster and consumes less GPU memory than ResNet101-dilated-os8.
Moreover, the overall computational efficiency of adding our {\it specialize and fuse} head upon the ResNet-decoder is still better than that of ResNet101-dilated-os8 (which is the optimal bound for many recent ResNet-based state-of-the-art methods).

\begin{table}[h]
    \centering
    \begin{tabular}{l|cc}
    \hline
    Method & \makecell{Testing\\FPS $\uparrow$} & \makecell{Training\\Memory $\downarrow$} \\
    \hline
    ResNet101                & 82 & 4.5G \\
    ResNet101-dilated-os8~\textdagger    & 24 & 9.6G \\
    ResNet101-decoder        & 32 & 8.4G \\
    ResNet101-decoder + our head & 26 & 8.9G \\
    \hline
    \multicolumn{3}{l}{\textdagger\footnotesize{Optimal bound for many recent works built upon dilated ResNet}}
    \end{tabular}
    \caption{
        Comparing the model efficiency measured on a GeForce RTX 2080 Ti with image size $512 \times 512$.
        FPS is averaged for processing $50$ images.
        GPU memory consumption in training is monitored with a batch size of $4$.
    }
    \label{table:speed_mem}
\end{table}

\section{Do different pyramid levels specialize in different classes?}
\begin{figure*}[h]
    \centering
    \includegraphics[width=0.95\textwidth]{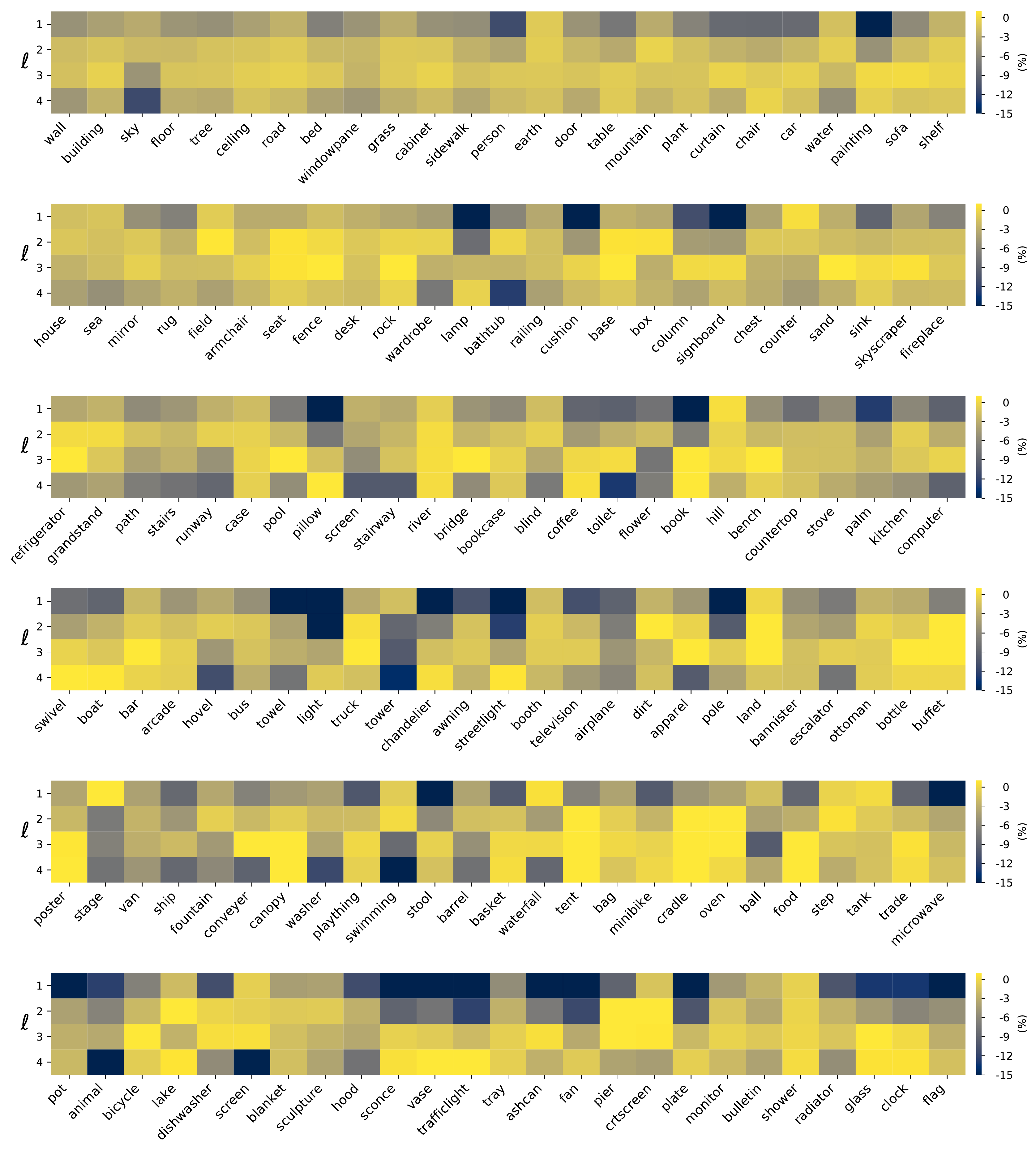}
    \caption{
    Semantic segmentation performance at each level $\{\hat{Y}^{(\ell)}\}_{\ell=1,\ldots,4}$ on different classes.
    We show the IoU difference between using prediction of a level $\hat{Y}^{(\ell)}$ and using the fused prediction $\hat{Y}$.
    A brighter cell indicates that the pyramid level is more specialized in the class.
    }
    \label{fig:detail_class}
\end{figure*}

To demonstrate our intuition that different pyramid levels have their specializations in different classes, we show the per-class IoUs predicted by each level of the semantic pyramid for the $150$ classes in the ADE20K dataset in Fig.~\ref{fig:detail_class}.
For clearer visualization, we show the IoU difference between the prediction of a single level $\hat{Y}^{(\ell)}$ and the final fused $\hat{Y}$ instead of the original IoU of $\hat{Y}^{(\ell)}$, and we clip the values in the heatmaps to the range from $-15\%$ to $1\%$.
A few entries in the heatmaps are larger than 0, which means that $\hat{Y}^{(\ell)}$ performs better than the fused $\hat{Y}$ on that class; it also implies that improving the performance of unity pyramid would provide opportunities for further enhancement.
In Fig.~\ref{fig:detail_class}, the coarser pyramid levels (\eg $\ell=1$) generally look inferior to the finer levels, which could be caused by the fact that the single-level evaluation setting disadvantage the coarser levels since the per-class IoUs for each level are evaluated on the same per-pixel scale.
However, this does not conflict with our intention to demonstrate that each pyramid level specializes in different classes.
In addition, the finest level performs significantly worst for some classes (\eg, {\sf sky, water, tower}) since it is not trained to predict the central parts of these large-area classes.
Some visual results presented in Fig.~\ref{fig:inter_112} provide more cues about this.

\section{Qualitative results} \label{sec:qual}
In Fig.~\ref{fig:qualitative}, we show some visual comparisons of results between our approach and a baseline method---ANL with HRNet48 backbone.
\begin{figure*}[h]
   \centering
    \begin{tabular}{c@{\hskip 1pt}c@{\hskip 1pt}c@{\hskip 1pt}c}
        input & ground truth & HRNet48-ANL & ours \\        \includegraphics[width=0.15\linewidth]{figs/qualitative/qual/ADE_val_00000014.jpg}&
        \includegraphics[width=0.15\linewidth]{figs/qualitative/qual/ADE_val_00000014.png}&
        \includegraphics[width=0.15\linewidth]{figs/qualitative/qual/ADE_val_00000014_pred_sem_anl.png}&
        \includegraphics[width=0.15\linewidth]{figs/qualitative/qual/ADE_val_00000014_pred_sem_final.png}\\
        \includegraphics[width=0.15\linewidth]{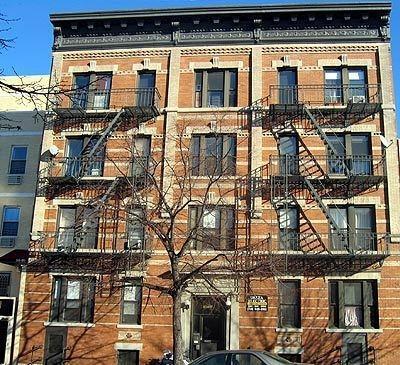}&
        \includegraphics[width=0.15\linewidth]{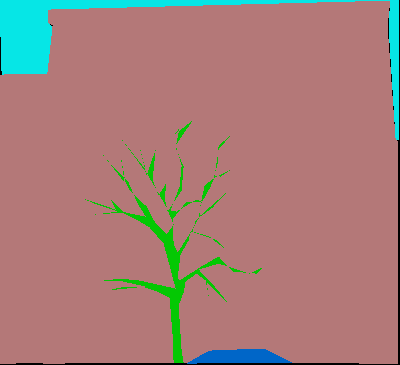}&
        \includegraphics[width=0.15\linewidth]{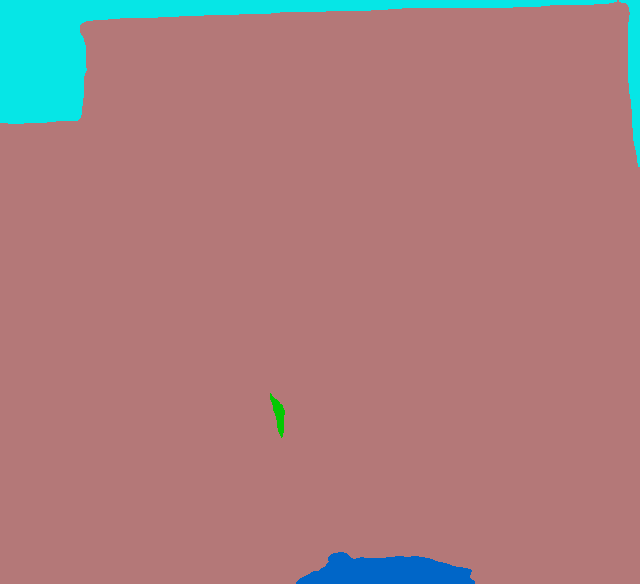}&
        \includegraphics[width=0.15\linewidth]{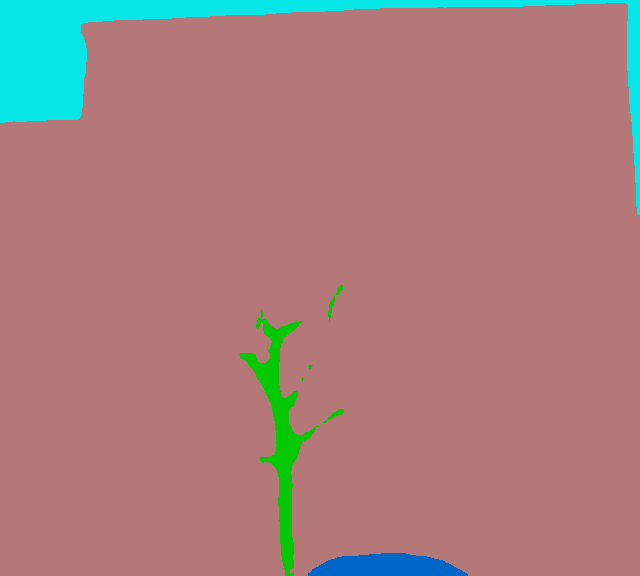}\\
        \includegraphics[width=0.15\linewidth]{figs/qualitative/qual/ADE_val_00000034.jpg}&
        \includegraphics[width=0.15\linewidth]{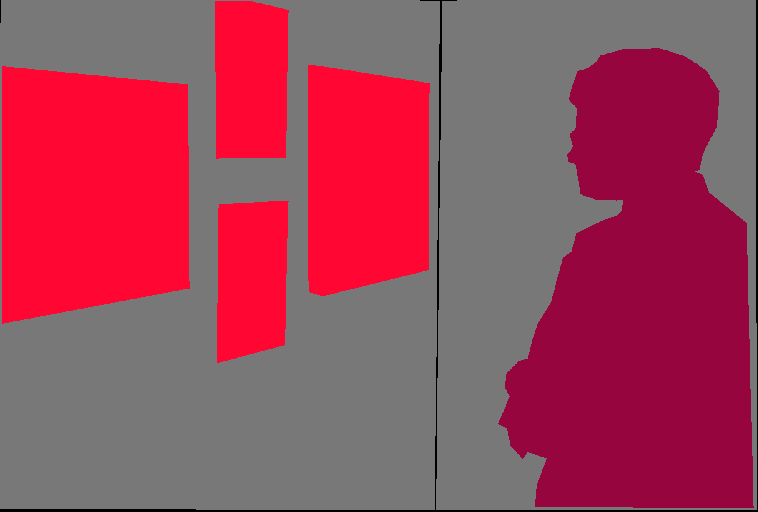}&
        \includegraphics[width=0.15\linewidth]{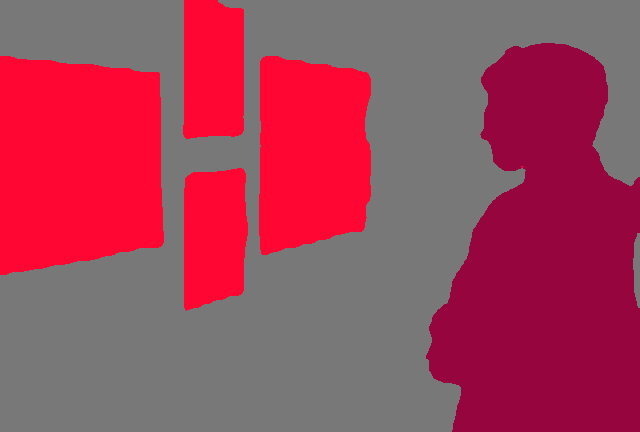}&
        \includegraphics[width=0.15\linewidth]{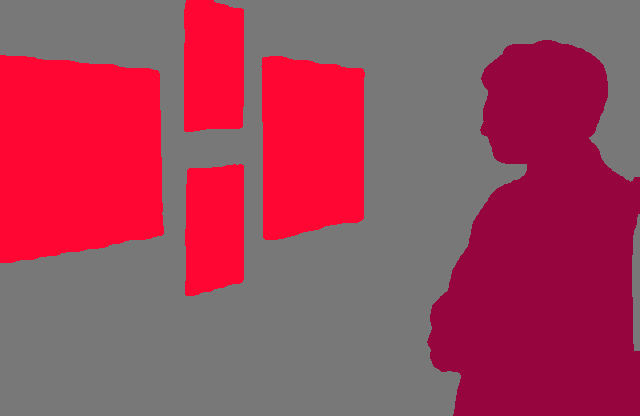}\\
        \includegraphics[width=0.15\linewidth]{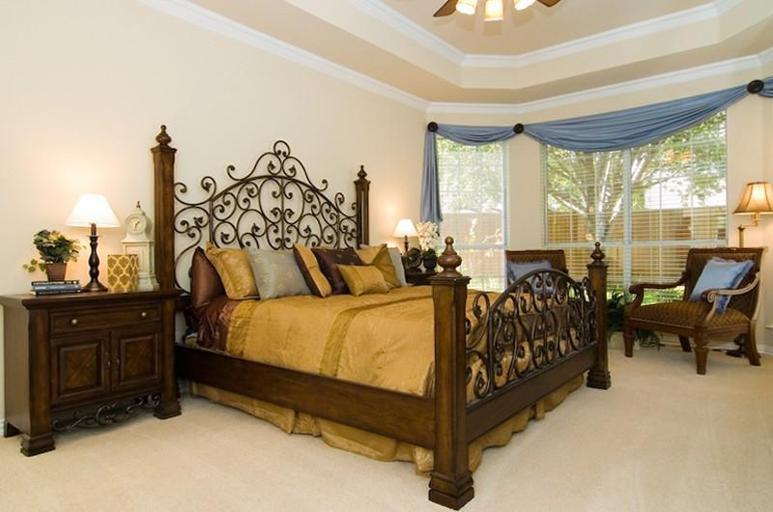}&
        \includegraphics[width=0.15\linewidth]{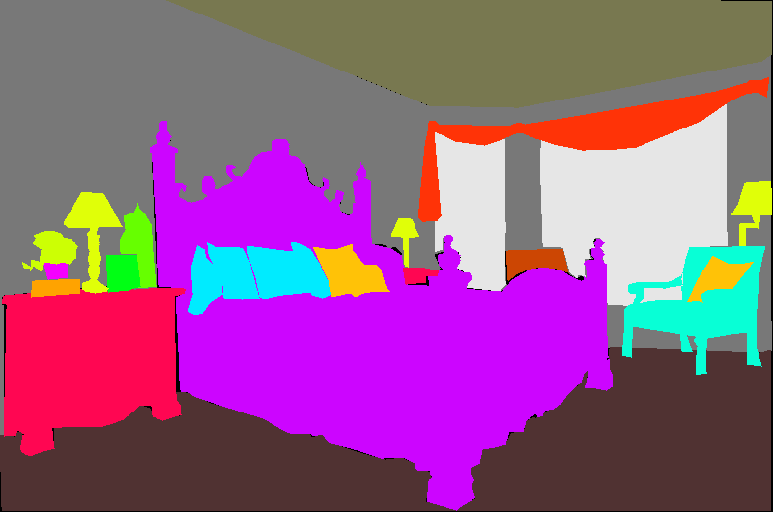}&
        \includegraphics[width=0.15\linewidth]{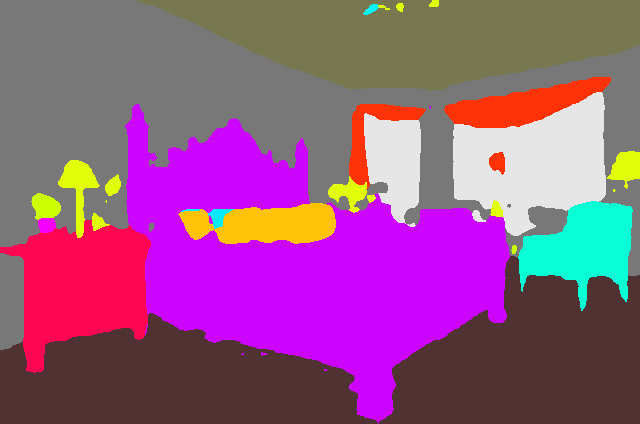}&
        \includegraphics[width=0.15\linewidth]{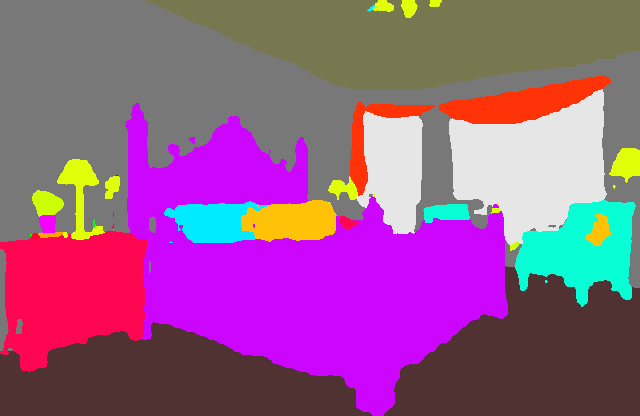}\\
        \includegraphics[width=0.15\linewidth]{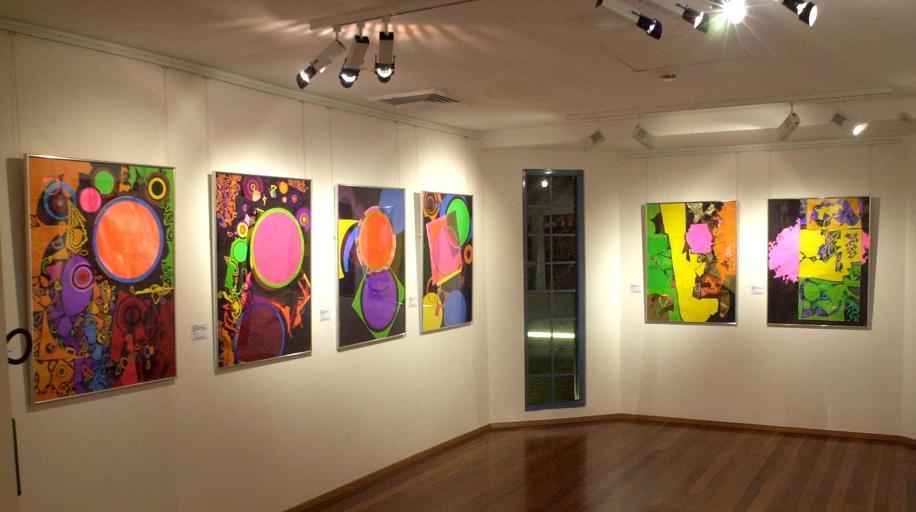}&
        \includegraphics[width=0.15\linewidth]{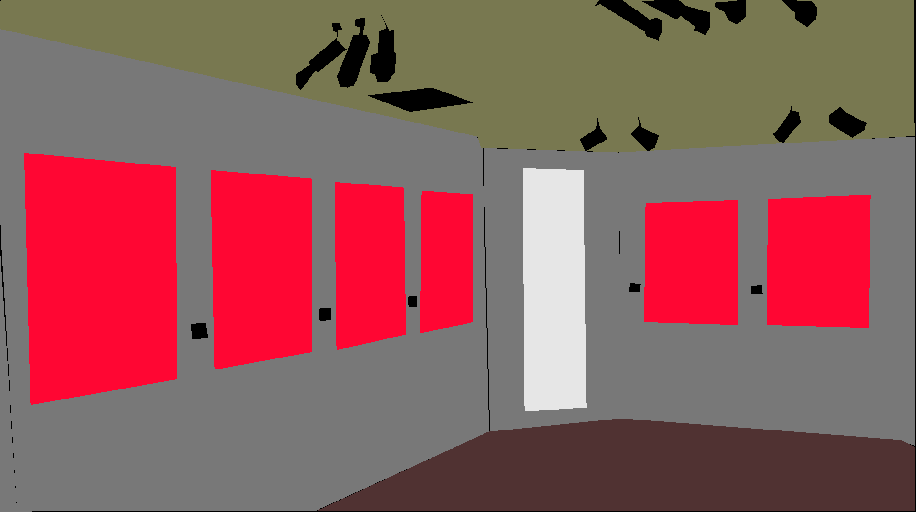}&
        \includegraphics[width=0.15\linewidth]{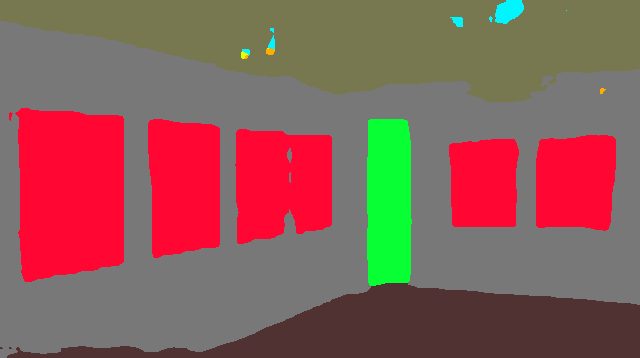}&
        \includegraphics[width=0.15\linewidth]{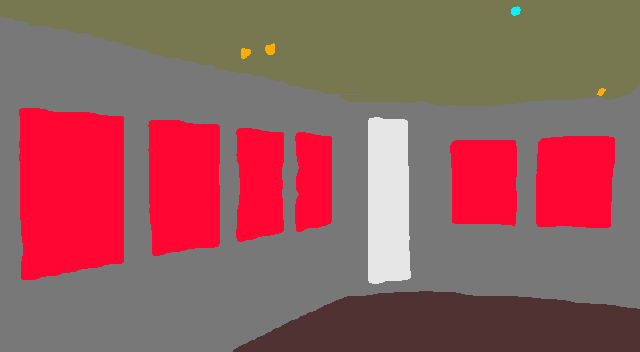}\\
        \includegraphics[width=0.15\linewidth]{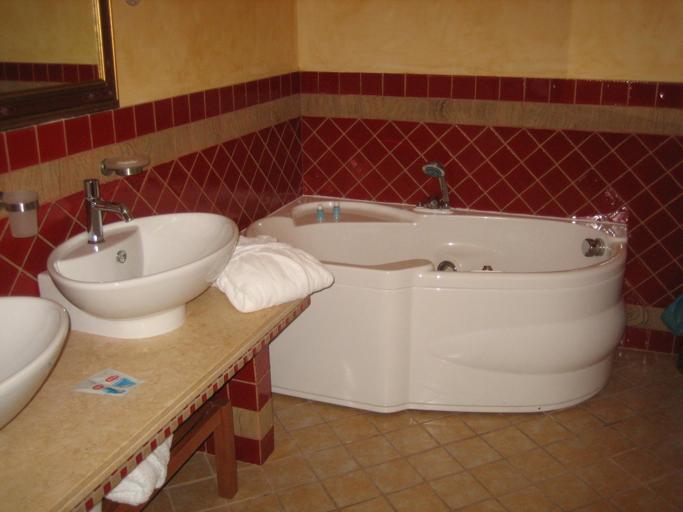}&
        \includegraphics[width=0.15\linewidth]{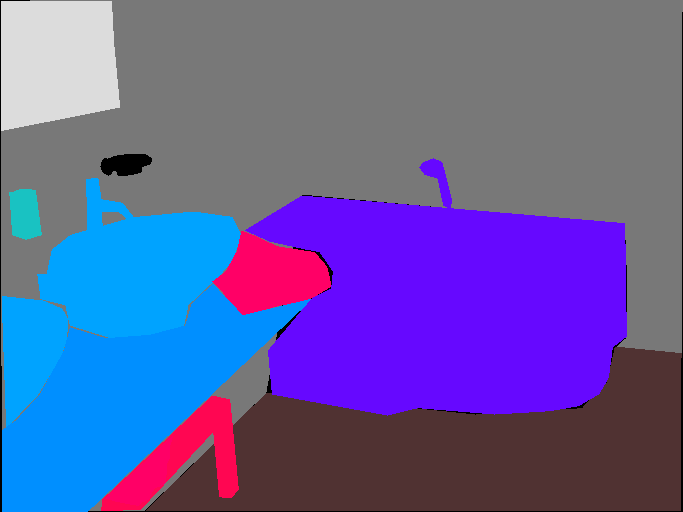}&
        \includegraphics[width=0.15\linewidth]{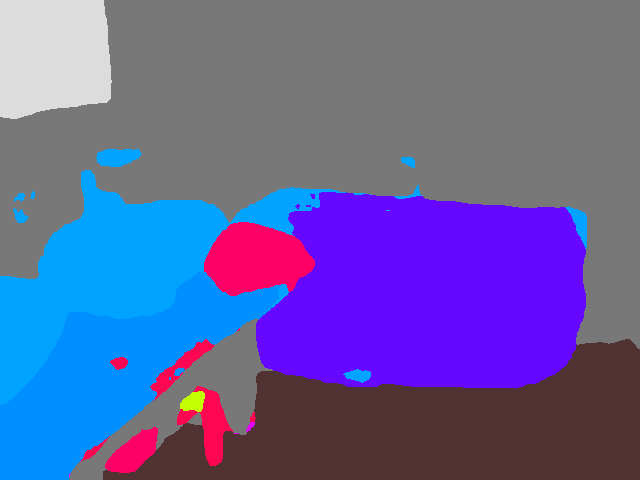}&
        \includegraphics[width=0.15\linewidth]{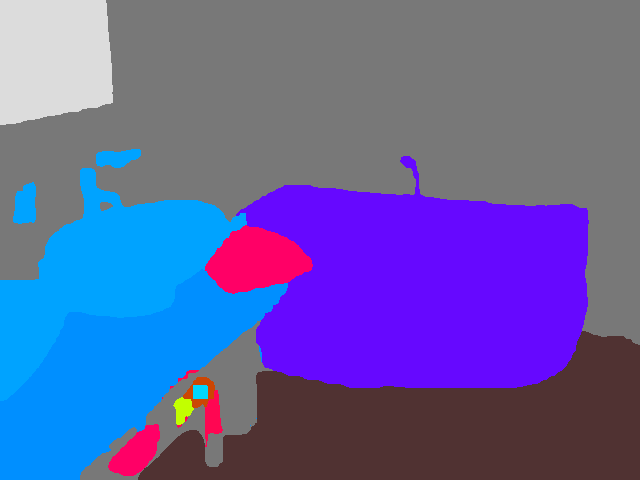}\\
        \includegraphics[width=0.15\linewidth]{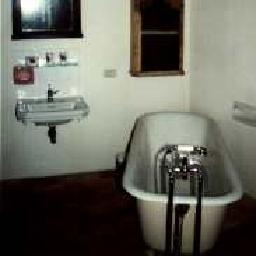}&
        \includegraphics[width=0.15\linewidth]{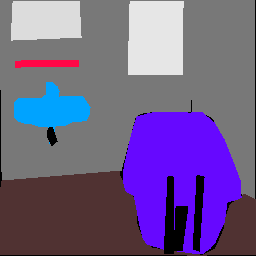}&
        \includegraphics[width=0.15\linewidth]{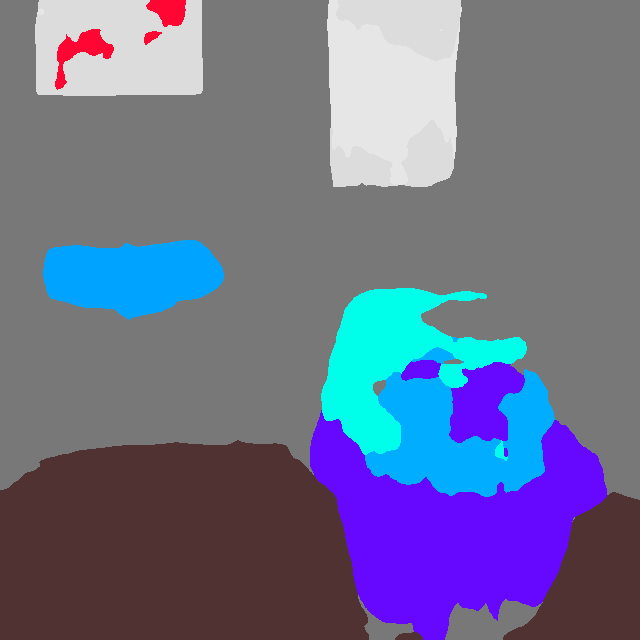}&
        \includegraphics[width=0.15\linewidth]{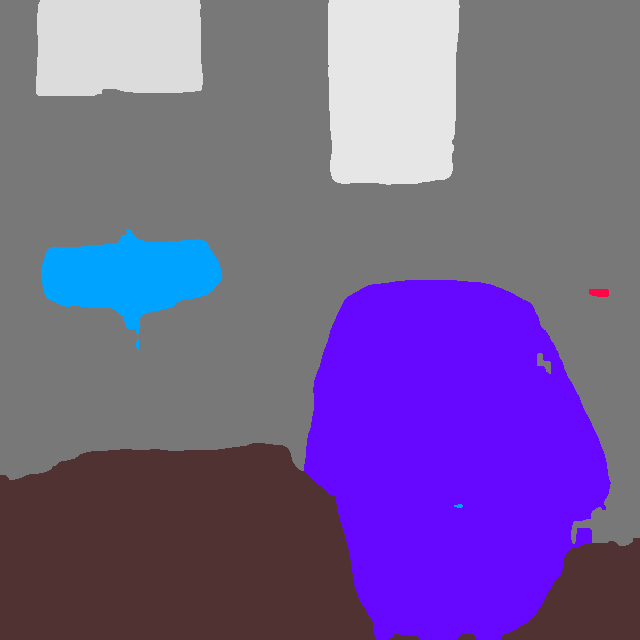}\\
        \includegraphics[width=0.15\linewidth]{figs/qualitative/qual/ADE_val_00000085.jpg}&
        \includegraphics[width=0.15\linewidth]{figs/qualitative/qual/ADE_val_00000085.png}&
        \includegraphics[width=0.15\linewidth]{figs/qualitative/qual/ADE_val_00000085_pred_sem_anl.png}&
        \includegraphics[width=0.15\linewidth]{figs/qualitative/qual/ADE_val_00000085_pred_sem_final.png}\\
        \includegraphics[width=0.15\linewidth]{figs/qualitative/qual/ADE_val_00000105.jpg}&
        \includegraphics[width=0.15\linewidth]{figs/qualitative/qual/ADE_val_00000105.png}&
        \includegraphics[width=0.15\linewidth]{figs/qualitative/qual/ADE_val_00000105_pred_sem_anl.png}&
        \includegraphics[width=0.15\linewidth]{figs/qualitative/qual/ADE_val_00000105_pred_sem_final.png}\\
        \includegraphics[width=0.15\linewidth]{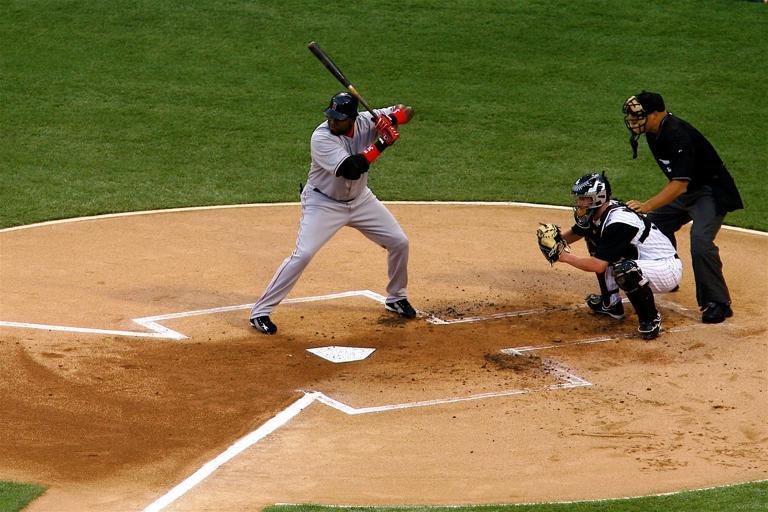}&
        \includegraphics[width=0.15\linewidth]{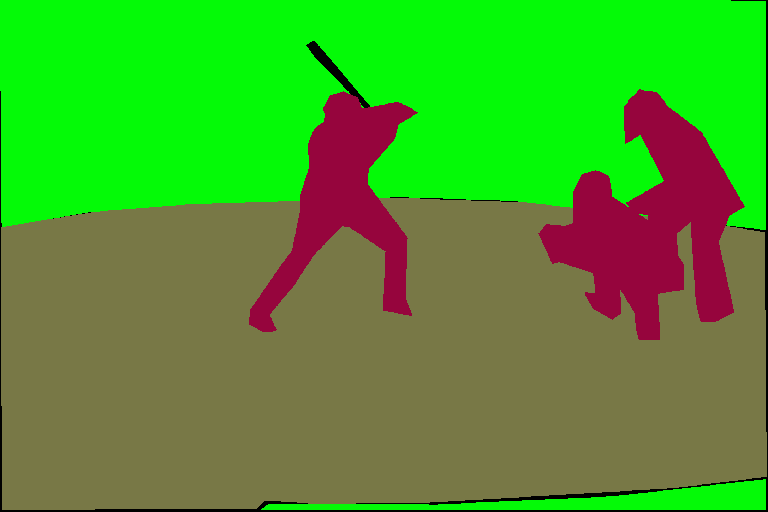}&
        \includegraphics[width=0.15\linewidth]{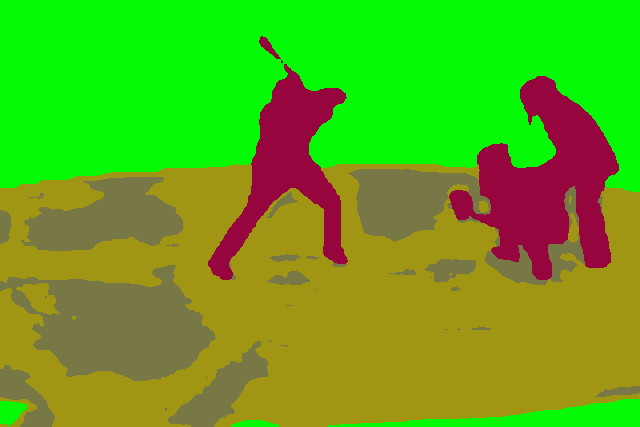}&
        \includegraphics[width=0.15\linewidth]{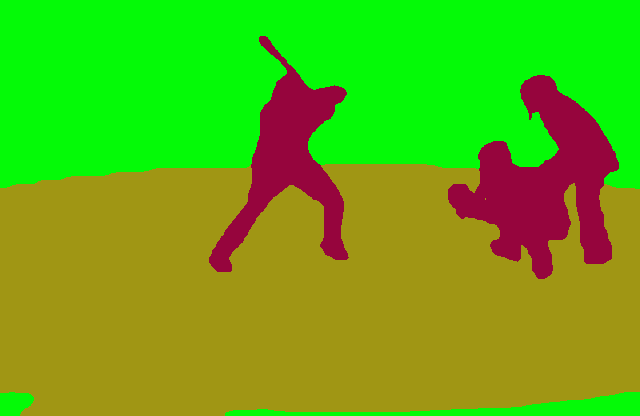}\\
    \end{tabular}
    \caption{Qualitative comparison. In the above examples, predictions by our approach are more consistent within an instance and also provide finer details.}
    \label{fig:qualitative}
\end{figure*}

\end{document}